\pdfoutput=1
\documentclass[runningheads]{llncs}
\usepackage[dvipsnames,table]{xcolor}
\usepackage{graphicx}

\usepackage{tikz}
\usepackage{comment}
\usepackage{amsmath,amssymb} 
\usepackage{color}
\usepackage[pagebackref,breaklinks,colorlinks]{hyperref}
\hypersetup{citecolor=[RGB]{119,185,0}}
\usepackage{tabularx,booktabs}
\usepackage{multirow}
\usepackage{pifont}
\usepackage{subfigure}
\usepackage{breakcites}

\usepackage[accsupp]{axessibility}  

\begin{document}
\pagestyle{headings}
\mainmatter
\def\ECCVSubNumber{1807}  

\title{ST-P3: End-to-end Vision-based Autonomous Driving via Spatial-Temporal Feature Learning} 

\titlerunning{ST-P3}
%
\author{Shengchao Hu\inst{1}$^\dagger$ \and
Li Chen\inst{2}$^\ast$ \and
Penghao Wu\inst{1,3}$^\dagger$ \and
Hongyang Li\inst{1,2} \and \\
Junchi Yan\inst{1,2} \and
Dacheng Tao\inst{4}
}
%
\authorrunning{S. Hu et al.}
%
\institute{MoE Key Lab of Artificial Intelligence, Shanghai Jiao Tong University \and
Shanghai AI Laboratory, Shanghai, China \and
The University of California, San Diego, CA, USA \and
JD Explore Academy, JD.com Inc., Beijing, China \\
\email{charles-hu@sjtu.edu.cn}\quad 
\email{lichen@pjlab.org.cn}
}
\maketitle

\let\thefootnote\relax\footnote{$^\ast$ Correspondence author. $^\dagger$ Work done during internship at Shanghai AI Laboratory.}

\begin{abstract}
Many existing autonomous driving paradigms involve a multi-stage discrete pipeline of tasks. To better predict the control signals and enhance user safety, an end-to-end approach that benefits from joint spatial-temporal feature learning is desirable. While there are some pioneering works on LiDAR-based input or implicit design, in this paper we formulate the problem in an interpretable vision-based setting. In particular, we propose a spatial-temporal feature learning scheme towards a set of more representative features for perception, prediction and planning tasks simultaneously, which is called ST-P3. Specifically, an egocentric-aligned accumulation technique is proposed to preserve geometry information in 3D space before the bird's eye view transformation for perception; a dual pathway modeling is devised to take past motion variations into account for future prediction; a temporal-based refinement unit is introduced to compensate for recognizing vision-based elements for planning. To the best of our knowledge, we are the first to systematically investigate each part of an interpretable end-to-end vision-based autonomous driving system. We benchmark our approach against previous state-of-the-arts on both open-loop nuScenes dataset as well as closed-loop CARLA simulation. The results show the effectiveness of our method. Source code, model and protocol details are made publicly available at \url{https://github.com/OpenPerceptionX/ST-P3}.

\end{abstract}

\section{Introduction}\label{sec: intro}

A classical paradigm design for autonomous driving systems often adopts a modular based spirit~\cite{behere2015functional,yurtsever2020survey},
where the input of a planning or controlling unit is based on the outputs from preceding modules in perception. 
As we witness the blossom of 
end-to-end algorithms and success applications into various domains~\cite{pfeiffer2017perception,hamalainen2019affordance}, there are some attempt implementing such a philosophy in autonomous driving as well~\cite{pomerleau1988alvinn,bojarski2016end,openpilot,bansal2018chauffeurnet,chitta2021neat,prakash2021multi,zhang2021end,chen2021wor,chen2022lav}. 
Rather than an isolated staged pipeline, we aim for a framework to directly take raw sensor data as inputs and generate the planning routes or control signals. 
A straightforward incentive to do so is that feature representations can thus be optimized simultaneously \textit{within} one network towards the ultimate goal of the system (\textit{e.g.}, acceleration, steering). 
%

One direction of end-to-end pipelines is to focus on the ultimate planning task mainly 
without explicit design of the intermediate representation~\cite{codevilla2018end,codevilla2019exploring,prakash2021multi,chen2020learning,chitta2021neat,hu2021safe}.
Reinforcement learning (RL) fits well as a feasible resolution since the planned routes are not unique and each action should be rewarded correspondingly based on the environment.
RL algorithms
are applied to mimic experienced human experts and guide the behavior learning of driving agents~\cite{zhang2021end,Chekroun2021gri}. 
Besides RL approaches, some propose to generate cost map with a trajectory sampler with knowledge of the perception environment~\cite{hu2021safe,chitta2021neat} or fusion of sensors in an attention manner~\cite{prakash2021multi,fadadu2022multi}.
These 
work aforementioned
achieve impressive performance on challenging scenarios in closed-loop simulation~\cite{dosovitskiy2017carla}. The plausible transfer from synthetic setting to realistic application remains an open question.

Another direction is to explicitly design the intermediate representations in the network, provide convincing interpretability of each module and thus enhance safety towards a stable and robust system. Based on the input
type, explicit approaches are divided into LiDAR-based
\cite{casas2021mp3,zeng2020dsdnet,zeng2019end,cui2021lookout}
and 
vision-based~\cite{hu2021fiery,philion2020lift,wang2021learning} 
respectively. 
LiDAR-based 
methods 
bundled with high-definition (HD) maps  exhibit good performance on various benchmarks, and they investigate the effect of each module of the system exhaustively. However, the inherent defects of LiDAR, such as recognition of 
traffic lights and short-range detection of objects, might confine their applications. 
%

\begin{figure}[tb!]
    \centering
    \includegraphics[width=1.\textwidth]{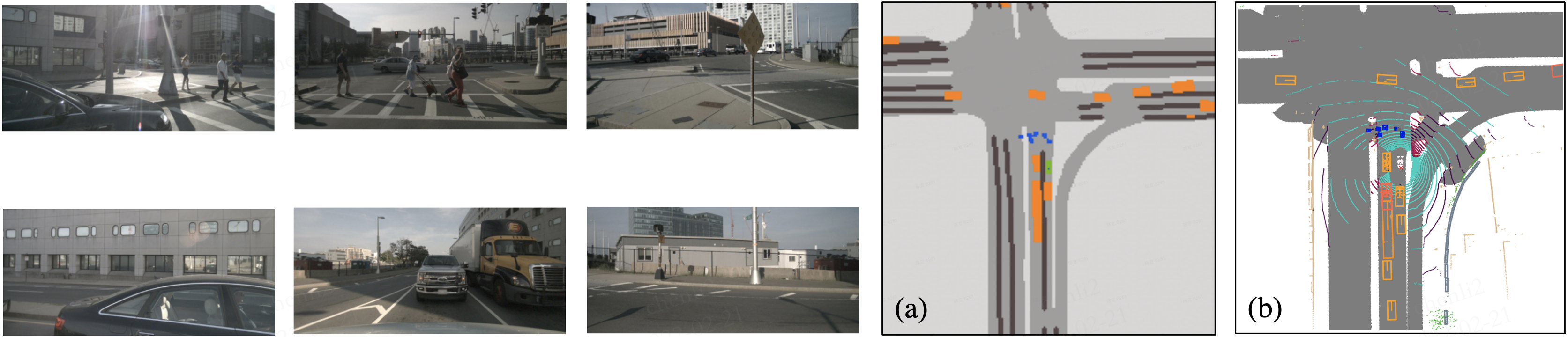}
    \caption{Problem setup whereby an interpretable vision-based end-to-end framework in (a) is devised, parallel to the LiDAR-based counterpart by aid of HD maps in (b)
    }
    \label{fig:motivation}
\end{figure}

In this paper, concurrent to the interpretable LiDAR-based pipelines~\cite{casas2021mp3,zeng2020dsdnet}, we propose to investigate the potential of \textit{vision-based} end-to-end framework (Fig. \ref{fig:motivation}).
If each module is exquisitely designed, to which extent the performance of each task as in perception, prediction and planning should be improved?

To answer the above question, the first key challenge for vision-based methods is to appropriately transform feature representations from  perspective views to the bird's eye view (BEV) space.  
The pioneering LSS approach~\cite{philion2020lift} extracts perspective features from multi-view cameras, lifts them into 3D with depth estimation and fuses into the BEV space. It is observed that latent depth prediction for the feature transformation between two views is crucial~\cite{hu2021fiery,wang2021learning,reading2021CaDDN}. On a theoretical analysis, this is true since lifting 2D planar information to 3D requires an additional dimension, which is the depth that fits into 3D geometric autonomous driving tasks. 
To further improve feature representation,
it is natural to incorporate temporal information into the framework as most scenarios are tasked with video sources.
Descending from~\cite{philion2020lift},
the follow-up literature project past frames' feature onto current coordinate view 
either by ego-motion of the self-driving vehicle (SDV) provided by dataset~\cite{hu2021fiery} or learned mapping from optical flow~\cite{wang2021learning}. 
These approaches project features in the past frame-by-frame in isolation and feed them into the temporal unit; instead
we accumulate all past aligned features 
in 3D space before transforming to BEV, preserving geometry information at best and compensating for more robust feature representations of the current state, which is empirically proven as a better design choice.





Equipped with representative features in BEV space, we formulate prediction task as the future instance segmentation, which is the same as does in FIERY~\cite{hu2021fiery}.
The common practice includes generating uncertainty from data distribution for current state and feeding it in a temporal model to infer predictions under a window of future horizons~\cite{hu2021fiery,wang2021learning}.
A natural incentive to boost 
future predictions, which is missing in current literature, is to take into account the motion variations in the \textit{past}. To do so, an additional temporal model with fusion unit could be introduced to reason about the probabilistic nature of both past and future motions. 
%
%
A stronger version of scene representations could therefore be obtained, which serves as recipe towards the ultimate planning task.


The general idea for motion planners is to output the most likely trajectory, given a sampling of possible candidates and semantic results from preceding modules~\cite{casas2021mp3,prakash2021multi,zeng2020dsdnet,hu2021safe}. In light of an interpretable spirit,
most previous work construct cost volumes, learning-based~\cite{casas2021mp3,chitta2021neat} and/or rule-based~\cite{zeng2019end,Chekroun2021gri,zeng2020dsdnet}, to indicate the confidence of trajectories with a certain form of trajectory modelling in a sampler.
We follow such a philosophy to indicate the most possible candidate with the help of a high-level command, without HD map as guidance. 
The outcome trajectory is further refined with the features from the front-view camera (front-view features)
to consider vision-based elements (\textit{e.g.}, traffic lights). This is inspired by MP3~\cite{casas2021mp3}, where they also remove HD maps and feed the network with a high-level command. However, we argue that the vision recognition module in~\cite{casas2021mp3} is off-the-shelve; in this work, we integrate vision information  in form of a lightweight GRU unit within the same network. Such a refinement process serves as a complementary feature boosting towards the final outcome.

\begin{figure}[t]
    \centering
    \includegraphics[width=0.98\textwidth]{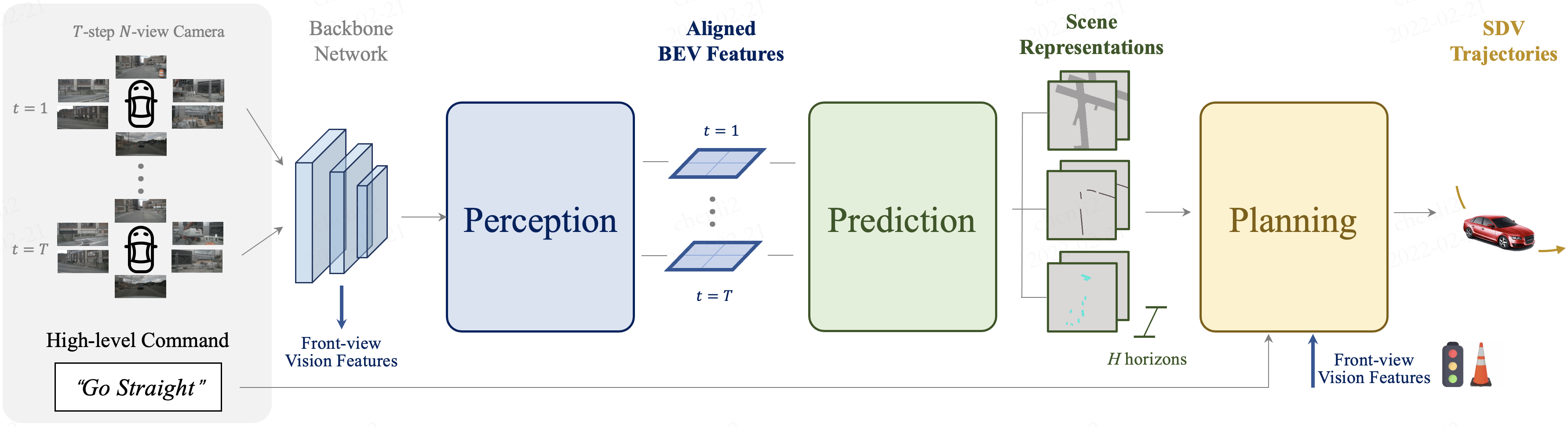}
    \caption{
    We present \textbf{ST-P3},  
    an interpretable end-to-end vision-based framework.
   For \textbf{perception}, 
   the egocentric aligned accumulation guarantees features (past and present) aligned and aggregated in 3D space
   to preserve geometry information before BEV transformation.
   For \textbf{prediction}, a dual pathway scheme is introduced to bring in past variations in pursuit of future predictions.
   For \textbf{planning}, the prior knowledge 
   is fed into a refinement unit to generate the  final trajectory with an integrated cost volume  and sampler from high-level commands
    }
    \label{fig:pipeline}
\end{figure}

To this end, we propose an interpretable vision-based end-to-end system that improves feature learning for perception, prediction and planning altogether, namely \textbf{ST-P3}. Fig. \ref{fig:pipeline} describes the overall framework. 
Specifically, given a set of surrounding camera videos, we feed them into the backbone to generate preliminary front-view features. An auxiliary depth estimation is performed to transform 2D features onto 3D space.
An egocentric aligned accumulation scheme first aligned past features to the current view coordinate system.
The current and past features are then aggregated in 3D space to preserve geometric information before the BEV representation. 
Apart from a commonly used temporal model for prediction,
this module is further boosted by constructing a second pathway to account for motion variations in the past. 
Such a dual pathway modelling ensures stronger feature representations to infer future semantic outcomes. 
Towards the ultimate goal of trajectory planning, we integrate prior knowledge 
from features in the early stage of the network. 
A refinement module is devised to generate the final trajectory with help of high-level commands and no presence of HD maps.
We benchmark our approach against previous state-of-the-arts on both open-loop nuScenes dataset as well as closed-loop simulator CARLA environment. 
To sum up, ST-P3 owns the following contributions baked into it:
\begin{enumerate}
    \item For better spatial-temporal feature learning, we propose three novel improvements, \textit{i.e.}, the egocentric aligned accumulation, the dual pathway modelling, and the prior-knowledge refinement for perception, prediction and planning modules respectively. The resulting new end-to-end vision-based network for autonomous driving is called ST-P3.
    %
    \item We investigate each part of an interpretable end-to-end system for autonomous driving tasks
    systematically. As a vision-based counterpart to the study of LiDAR-based approaches~\cite{zeng2020dsdnet}, to our best knowledge, we provide the first detailed analysis and comparison for a vision-based pipeline.
    \item ST-P3 achieves state-of-the-art performance on benchmarks from the popular nuScenes dataset and simulator CARLA. The full suite of codebase, as well as protocols are made publicly available.
\end{enumerate}



\section{Related Work}\label{sec: related work}
We briefly discuss the related works in four aspects.

\noindent\textbf{Interpretable End-to-End Framework. }
We review popular approaches~\cite{cui2021lookout,zeng2019end,zeng2020dsdnet,sadat2020perceive,casas2021mp3} that adopt an explicit design to have clear interpretability of the system and hence prompts safety.
These are LiDAR-based approaches mainly as there are few vision-based solutions in the wild (compared in the previous section already).
%
The pioneering NMP~\cite{zeng2019end} takes as input LiDAR and HD maps to first 
predict the bounding boxes of actors in the future, and then learns a cost volume to choose the best planned trajectory.
The subsequent P3 
\cite{sadat2020perceive} work
further achieves consistency between planning and perception 
by
a differentiable occupancy representation, which
is explicitly used as cost by planning to produce safe maneuvers. 
To avoid heavy reliance on HD maps, 
Casas \textit{et al.} (MP3)
\cite{casas2021mp3} 
constructs an online map from segmentation as well as the current and future states of other agents;
%
these 
results are then fed into 
a sampler-based planner to obtain a safe and comfortable trajectory. 
LookOut~\cite{cui2021lookout}
predicts a diverse
set of futures of how the scene might unroll and estimates
the trajectory by optimizing a set of contingency
plans.
%
DSDNet~\cite{zeng2020dsdnet} 
considers the interactions between actors and produces socially consistent multimodal future predictions. 
It explicitly exploits the predicted
future distributions of actors to plan a safe maneuver by using a structured planning cost.
In general, LiDAR based approaches demonstrate good performance on challenging urban scenarios.
%
Unfortunately, the datasets and baselines in these work are not released 
to compare.

%


%

\noindent\textbf{Bird's eye view (BEV) Representation}
is a natural and perfect fit for planning and control tasks~\cite{ng2020bev,zhang2021end,9123682,chitta2021neat,bansal2018chauffeurnet},
since it avoids problems such as occlusion, scale distortion, \textit{etc.}, and preserves 3D scene layout. 
Although information in LiDAR and HD maps can be easily represented in BEV, 
how to project vision inputs from camera view to BEV space 
is a non-trivial problem.
Some learning based methods~\cite{chitta2021neat,9123682} implicitly project  image input into BEV, but the quality can not be guaranteed since usually we do not have ground truth in BEV to supervise the projection process. 
%
Loukkal \textit{et al.}
\cite{loukkal2021driving} explicitly projects image into BEV using homography between image and BEV plane.
\cite{li2022bevformer,chen2022persformer} aquires BEV features through spatial cross-attention with pre-defined BEV queries.
LSS~\cite{philion2020lift} and FIERY~\cite{hu2021fiery} perform the projection
with estimated depth and image intrinsic, which have shown impressive performance. 
We predict depth and project images 
in a similar fashion. 
Different from FIERY~\cite{hu2021fiery} which transforms past features to current timestamp frame-by-frame correspondingly,
we append all past 3D features to the current ego view (egocentric)  
and accumulate the aligned features, providing better representation for subsequent tasks.
%

\noindent\textbf{Future Prediction.}
Current motion prediction methods~\cite{ngiam2021scene,gu2021densetnt,pmlr-v164-jia22a,LiaoICME22} usually takes ground truth perception information and HD map as input, but they are susceptible to cumulative error when the perception input comes from other modules in real-life application. Taking raw sensor data as input,  end-to-end methods which focus on future trajectory prediction or take it as an intermediate step usually rely on LiDAR and HD map~\cite{luo2018fast,zeng2020dsdnet,zeng2019end,cui2021lookout,wei2021perceive} to detect and predict.
Recently, future prediction in the form of BEV semantic segmentation using only camera input~\cite{hu2021fiery,wang2021learning} has been proposed and shown great performance. However, the evolution process of the past is not well captured and exploited~\cite{hu2021fiery,philion2020lift,wang2021learning}. Inspired by video future prediction~\cite{Hu2020ProbabilisticFP}, we combine the probabilistic  uncertainty with the dynamics of past to predict diverse and plausible future scenes.

\noindent\textbf{Motion Planning.}
We cover previous learning-based motion planning methods and omit traditional approaches in this part. 
For implicit methods~\cite{pomerleau1988alvinn,codevilla2019exploring,prakash2021multi}, the network directly generates planned trajectory or control commands. Although such design is direct and simple, it suffers from robustness and lack of interpretability. On the contrary, explicit methods usually 
build a 
cost map with a trajectory sampler 
to generate the desired trajectory by choosing the optimal candidate with the lowest cost. 
The cost map can be constructed with hand-crafted rules~\cite{sadat2020perceive,casas2021mp3,cui2021lookout} based on intermediate representations such as  
segmentations and HD maps; 
or it can be learned directly from the network~\cite{zeng2019end}. DSDNet~\cite{zeng2020dsdnet} combines hand-crafted and learning-based costs to obtain an integrated cost volume. We also adopt this combination 
to choose the best trajectory. However, we modify the pipeline by adding an additional GRU refinement unit 
with navigation signal to further adjust and optimize the chosen trajectory.

\section{Methodology}\label{sec: method}

An overview of ST-P3 is given in Fig.~\ref{fig:pipeline}.
ST-P3 first extracts features of a sequence of surrounding images and lift them to 3D with depth prediction. They are fused in both spatial and temporal domains with the  egocentric aligned accumulation
(see Sec.~\ref{sec: method-perception}).
We show the dual pathway mechanism in Sec.~\ref{sec: method-prediction}, a novel uncertainty modeling to incorporate history information.
Sec.~\ref{sec: method-plan} elaborates on how we  utilize the intermediate representations to plan a safe and comfortable trajectory.

\subsection{Perception:  Egocentric Aligned Accumulation}
\label{sec: method-perception}

\begin{figure}[t]
    \centering
    \begin{minipage}[b]{0.59\textwidth}
      \includegraphics[width=\textwidth]{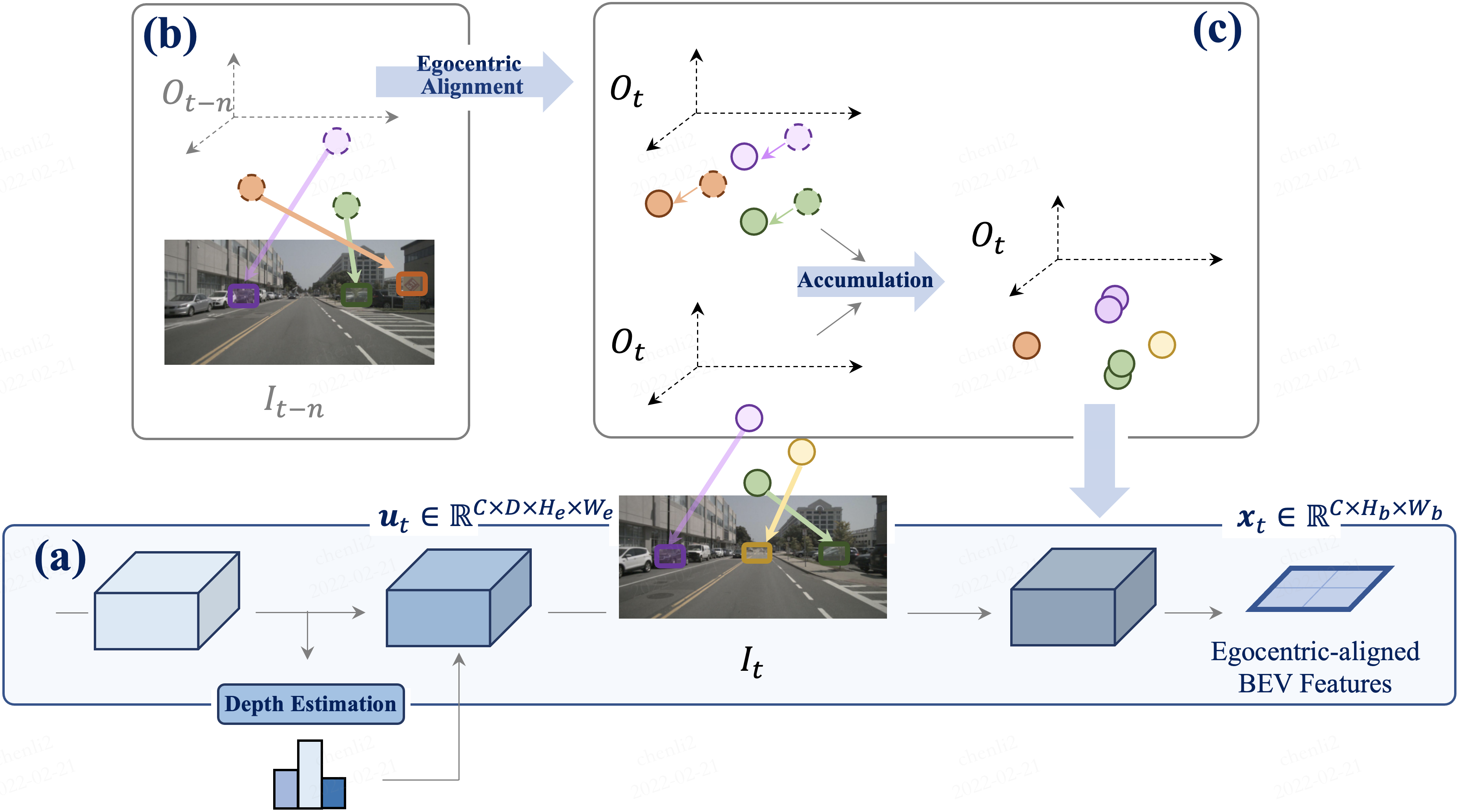}
    \end{minipage} ~~ 
    \begin{minipage}[b]{0.37\textwidth}
      \caption{
      Egocentric aligned accumulation for Perception. \textbf{(a)} Feature at current timestamp is lifted to 3D with depth estimation, and pooled to BEV features $x_t$ after alignment; 
      \textbf{(b-c)}
      3D features in previous frames are aligned to current 
      view and fused 
      with all past and current states, so that 
      the feature representation 
      get enhanced
      }
    \label{fig:perception}
    \end{minipage}
\end{figure}

In this stage we need to build a spatiotemporal BEV feature from multi-view camera inputs in past $t$ timestamps.
As discussed in Sec.~\ref{sec: intro}, the direct concatenation~\cite{wang2021learning} has the alignment issue while FIERY~\cite{hu2021fiery} suffers from losing height information.
Towards these problems, here we introduce our accumulative ego-centric alignment method which incorporates two steps, \textit{i.e.}, spatial and temporal fusion.
In the spatial fusion part, multi-view images in all timestamps are processed and transformed to the current ego-centric 3D space.
While in the temporal fusion step, we enhance the feature discrimination of static elements and objects in motion in an accumulative way and adopt a temporal model to achieve the final fusion.
An illustration is depicted in Fig.~\ref{fig:perception}.

\noindent\textbf{Spatial Fusion.}
On one hand, features from multi-view images should be transformed to a common frame.
Inspired by~\cite{philion2020lift}, we extract the feature and predict the corresponding depth for each image and then lift it into the global 3D frame. 
In particular, each camera image $I_t^n \in \mathbb{R}^{3 \times H \times W}$ is passed through a backbone network to obtain features $f_i^k \in \mathbb{R}^{C \times H_e \times W_e}$ and 
depth estimation $d_i^k \in \mathbb{R}^{D \times H_e \times W_e}$ respectively, 
where $i \in \{1,\dots,t\}$, $n \in \{1,\dots,6\}$, $C$ is the number of feature channels, $D$ denotes the number of discrete depths and $(H_e, W_e)$ indicates the spatial size.
%
%
%
%
Since the exact depth information is not available, we spread the feature across the entire ray of space according to the predicted depth distribution by taking the outer product of the matrices:
\begin{equation}
	u_i^k = f_i^k \otimes d_i^k,
\end{equation}
with $u_i^k \in \mathbb{R}^{C \times D \times H_e \times W_e}$.
%
%
Then we use the camera extrinsics and intrinsics to transform the camera feature frustums $u_i \in \{u_i^1,\dots,u_i^n\}$ to the global 3D coordinate whose origin is at the inertial center of the ego-vehicle at time $i$.
%

On the other hand, the spatial fusion needs to align past features to the current frame for the downstream prediction and planning tasks.
With the ego-motion from time $t-1$ to $t$, we can transform the cube obtained at time $t-1$ into the coordinate system centered on the SDV at time $t$.
The same process could be applied to the past frames as well, resulting in ego-centric features $\{u_i^{'}\}$ for all previous timestamps.
%
Ultimately, the BEV feature maps $b_i \in \mathbb{R}^{C \times H_b \times W_b}$ could be sum pooled from $\{u_i^{'}\}$ along the vertical dimension.

\noindent\textbf{Temporal Fusion.}
Classical temporal fusion methods directly exploit 3D convolutions with stacked BEV features.
However, it is noted that 
the corresponding features at the same location of 
various cubes should be similar if there are objects that are stationary on the ground, such as lanes and static vehicles.
Due to this property,
we 
perform
a self-attention to boost the perceptual ability of static objects by adding the previous BEV feature maps to the current, which can be formulated as follows (where the discount $\alpha = 0.5$ and $\tilde{x}_1 = b_1$): 
%
\begin{equation}
\tilde{x}_t = b_t + \sum_{i=1}^{t-1} \alpha^i \times \tilde{x}_{t-i},
\end{equation}

In order to perceive dynamic objects more accurately, we then feed these features into a temporal fusion network achieved by 3D convolution. To compensate for deviations caused by ego-vehicle motion, we add the motion matrix to the features by concatenating it in the spatial channel:
\begin{equation}
    x_{1\sim t} = \mathcal{C}(\tilde{x}_{1 \sim t}, m_{1 \sim t}),
\end{equation}
where $m_{1 \sim t}$ denotes the ego-motion matrices and $\mathcal{C}$ denotes the 3D conv network.


\subsection{Prediction: Dual Pathway Probabilistic Future Modelling}\label{sec: method-prediction}

In a dynamic driving environment, traditional motion prediction algorithms~\cite{wu2020motionnet,fang2020tpnet,djuric2020uncertainty} often predict future trajectories as a deterministic or multi-modal results.
%
However, a finite number of probabilities modelling could not cover the complex situation of future due to the interaction among agents, traffic elements and road environments.
In order to deal with the stochasticity of the future, we wish to reason about the conditional uncertainty in the prediction features.
Motivated by Hu \textit{et al}.~\cite{hu2020probabilistic}, we model the future uncertainty as diagonal Gaussians with mean $\mu \in \mathbb{R}^L$ and variance $\sigma^2 \in \mathbb{R}^L$, where $L$ is the latent channel.
Samples from the distribution could serve as a hidden state feature for future use.
It is noted that during training, we use samples $\eta_t \sim \mathcal{N}(\mu_{\mathrm{t}}, \sigma^2_{\mathrm{t}})$ while it will be sampled from $\eta_t \sim \mathcal{N}(\mu_{\mathrm{t}}, 0)$ during inference time.

\begin{figure}[t]
    \centering
    \begin{minipage}[b]{0.59\textwidth}
      \includegraphics[width=\textwidth]{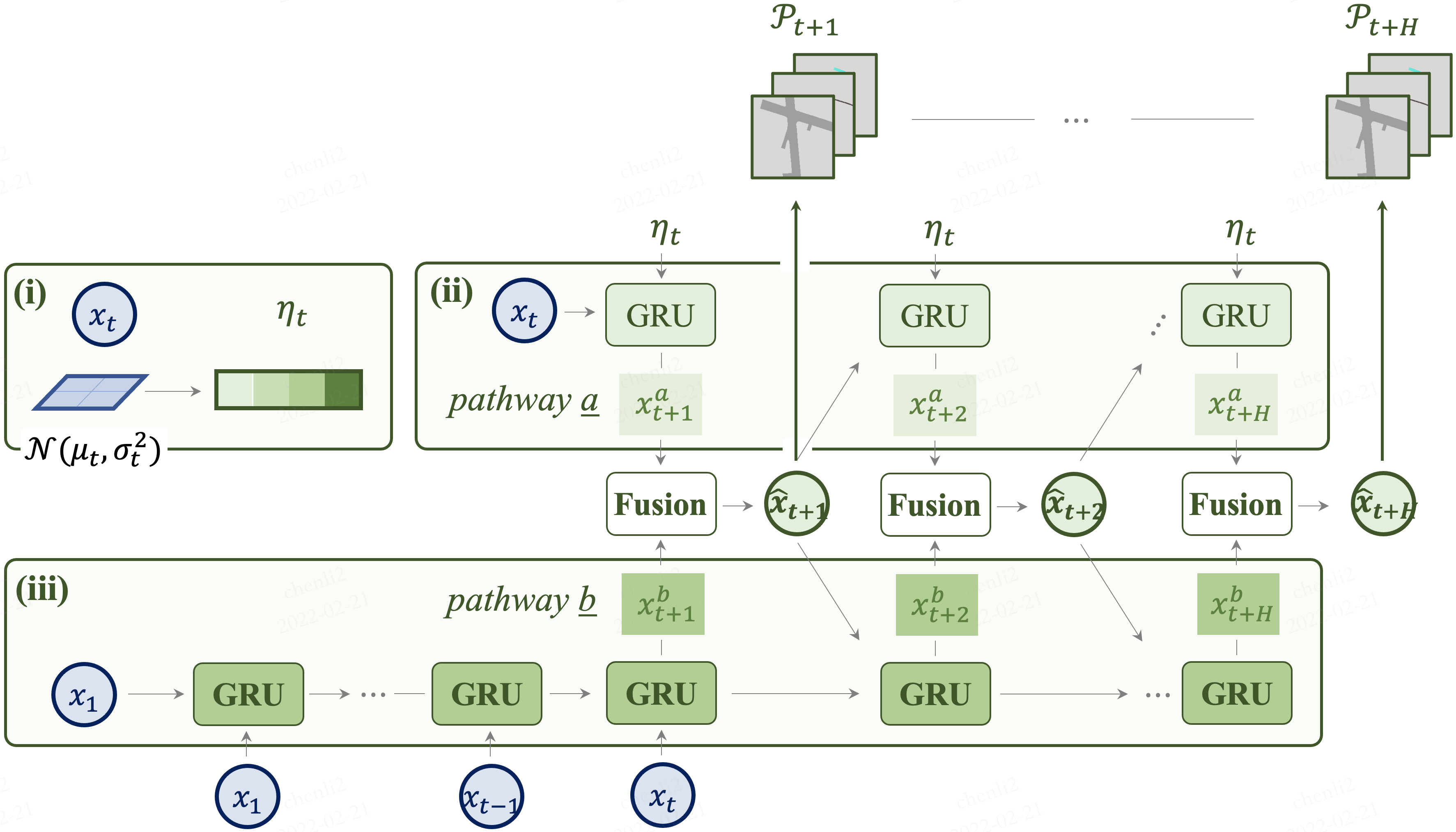}
    \end{minipage} ~~ 
    \begin{minipage}[b]{0.37\textwidth}
      \caption{
      Dual pathway modelling for Prediction.
      \textbf{(i)} the latent code is from distribution of feature maps; \textbf{(ii-iii)}
      pathway \underline{\textit{a}} incorporates uncertainty distribution to indicate the multi-modal of future, 
      while  
      \underline{\textit{b}} learns from  variations in the past, which is beneficial to compensate for information in \underline{\textit{a}}
      }
    \label{fig:prediction}
    \end{minipage}
\end{figure}


%
%
The architecture of the prediction module 
is depicted in Fig.~\ref{fig:prediction}.
We integrate the BEV features till current timestamp and the future uncertainty distribution into our prediction model, corresponding to two pathways in the dual modelling respectively.
%
%
One uses historical features $(x_1, \dots, x_t)$ as GRU inputs, and $x_1$ as the initial hidden state for prediction. The other uses samples from $\eta_t$ as the GRU input and $x_t$ as the initial hidden state. 
When predicting the feature at time $t+1$, we combine the predicted features in the form of a mixed Gaussian:
\begin{equation}
    \hat{x}_{t+1} = \mathcal{G}(x_t, \eta_t) \oplus \mathcal{G}(x_{0:t}),
\end{equation}
where $\mathcal{G}$ represents the process of GRU.
And we use the mixture prediction as the base for the following prediction progress.
Through this method, Dual Modelling recursively predicts future states $(\hat{x}_{t+1},\dots, \hat{x}_{t+H})$.

\begin{figure}[tb!]
    \centering
    \includegraphics[width=\textwidth]{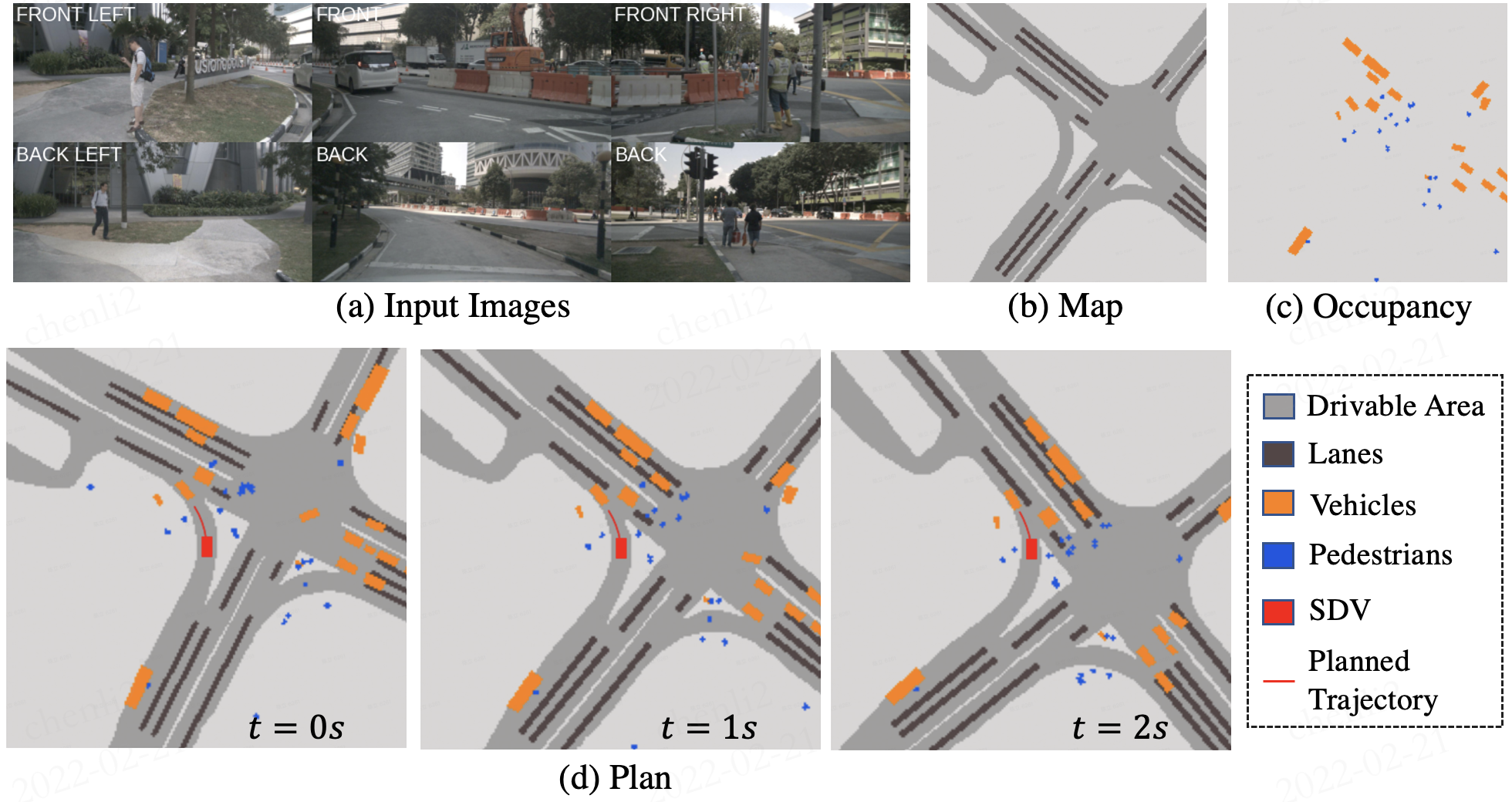}
    \caption{Qualitative results on nuScenes. (b) shows predicted map representation including drivable area and lanes. (c) depicts the segmentation of vehicles and pedestrians. (a-c) are at $t=1s$. (d) represents the overall results from our model - perception, prediction, planning. The intermediate scene representations are robust in the time period, and the SDV successfully generate a safe trajectory to do left-turn without collision with curbsides or the front car
    }
    \label{fig:pred outputs}
\end{figure}

All the features $(x_1, \dots, x_t), (\hat{x}_{t+1},\dots, \hat{x}_{t+H})$ are fed into the decoder $\mathcal{D}$ which has multiple output heads to generate different interpretable intermediate representations. 
The outcome is shown in Fig.~\ref{fig:pred outputs}.
For the instance segmentation task, we follow the evaluation metrics in~\cite{hu2021fiery}, where the heads output the instance centerness, offset and future flow.
Meanwhile, the semantic segmentation head mainly focuses on the vehicle and pedestrian which are the main actors in a driving setting.
Furthermore, 
as HD map plays a vital role in autonomous driving~\cite{sadat2020perceive,sadat2019jointly,zeng2019end}, we explicitly generate two elements - drivable area and lanes, to provide an interpretable map representation.
%
%
A cost volume head is designed for representing the \textit{expense} of each possible location that the SDV can take within the planning horizon. More detailed information on cost volume is illustrated in Sec.~\ref{sec: method-plan}.
Note that we also decode features in the past frames to boost historical features' accuracy, which is required by Dual Modelling. As demonstrated in Sec.~\ref{sec: res-nuscenes}, more accurate feature information could lead to better prediction.

\subsection{Planning: Prior Knowledge Incorporation and  Refinement}
\label{sec: method-plan}


\begin{figure}[t]
    \centering
    \begin{minipage}[b]{0.6\textwidth}
      \includegraphics[width=\textwidth]{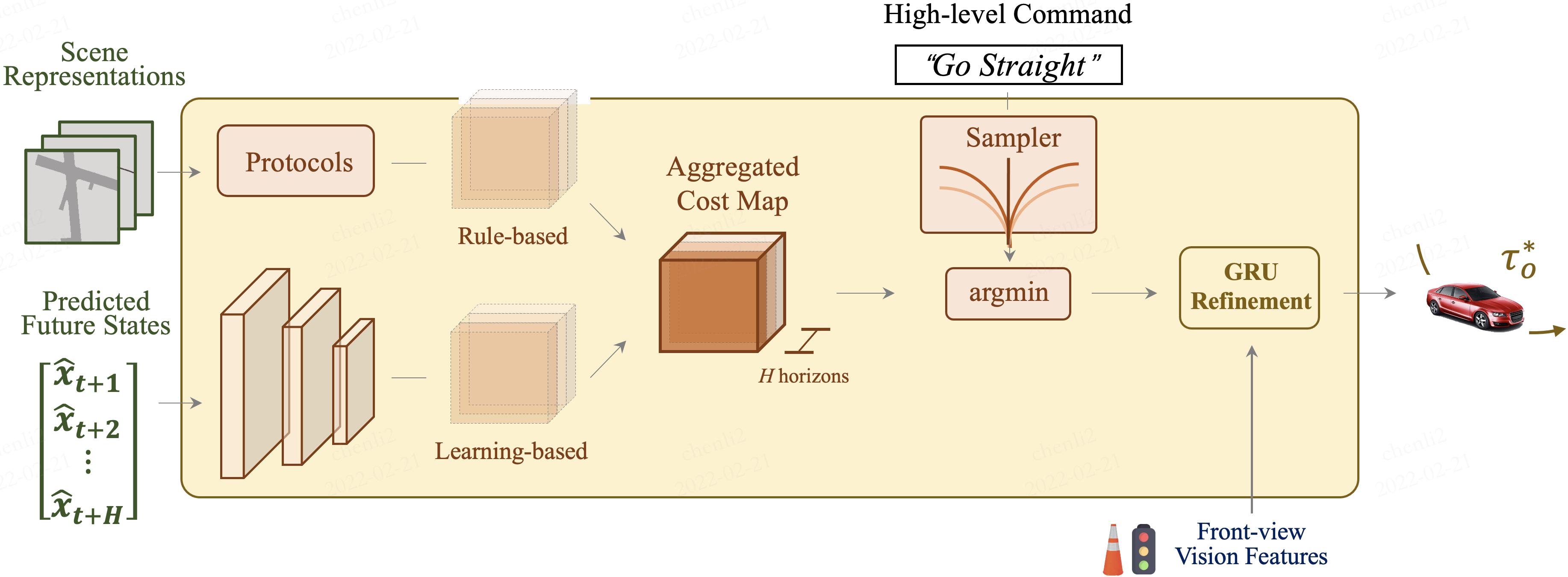}
    \end{minipage}  
    \begin{minipage}[b]{0.37\textwidth}
      \caption{
      Prior knowledge integration and refinement for Planning. 
      The overall cost map includes two subcosts. 
      The min-cost trajectory is further refined with front-view features to aggregate vision-based information from camera inputs
      }
    \label{fig:planning}
    \end{minipage}
\end{figure}

As the ultimate goal, a motion planner needs to plan a safe and comfortable trajectory towards the target point.
Given the occupancy predictions $o$ and perceptions of map representations $m$, we design the motion planner which samples a diverse set of trajectories, and picks the one minimizing the learned cost function, inspired by~\cite{zeng2019end,sadat2020perceive,casas2021mp3}.
However, we differentiate from them with an additional optimization step with a temporal model to integrate information of target point and traffic lights.
The overall module is illustrated in Fig.~\ref{fig:planning}.

The cost function makes full use of the learned occupancy probability field (segmentation maps in Prediction) and rich pre-knowledge as well, such as traffic rules, to ensure the safety and smoothness of the final trajectory.
Formally, given the SDV's dynamic state, we adopt the bicycle model~\cite{polack2017kinematic} to sample a set of trajectories $\tau$.
%
%
The objective cost function $f$ is composed of subcosts, $f_o$ that evaluates the predicted trajectory at every timestamp $t$ according to the prior knowledge, $f_v$ from prediction decoder (learning-based), and $f_r$ that relates to the overall performances of the trajectory such as the comfort, progress.
Thus the overall cost function can be an equally weighted combination as:
\begin{equation}
    f(\tau, o, m; w) = f_o(\tau, o, m; w_o) + f_v(\tau ; w_v) +f_r(\tau; w_r),
\end{equation}
with $w = (w_o,w_v,w_r)$ being the vector of all learnable parameters.
We briefly describe the subcosts below and refer readers to the Appendix for details.

\noindent\textbf{Safety Cost.} The SDV should not collide with other objects on the road and need to consider their future motion when planning its trajectory.
The planned trajectories cannot overlap the grids occupied by other agents or road elements and need to maintain a certain safe distance at the high-velocity motion.
%

\noindent\textbf{Cost Volume.} Due to the complexity of road information, we cannot manually enumerate all possible cases or planning cost, thus we introduce a learned Cost Volume generated by the head in Sec.~\ref{sec: method-prediction}.
In order to balance the cost scale, we clip the value so that it does not take a dominant role in evaluating trajectories.

\noindent\textbf{Comfort and Progress.} We penalize trajectories with large lateral acceleration, jerk or curvature.
We wish the trajectory to be efficient towards the destination, hence the trajectory progressing forward would be awarded.

However, the above cost does not contain the target information which is often provided by a routed map; yet, this is not available in our setting.
Thus we adopt the high level command including forward, turn left and turn right, and only evaluate the trajectories according to the corresponding command. 
%
%
%
To sum up, the motion planner now outputs one with the minimum cost as:
\begin{equation}
    \tau^* = \arg\min_{\tau_h} f(\tau_h, c) = \arg\min_{\tau_h} f(\tau_h, o, m; w), 
    \label{eq:over_obj}
\end{equation}
where $\tau^*$ is the selected trajectory, $\tau_h$ is the trajectory set under the corresponding high level command and $c$ is the overall cost map.
%
%
Furthermore, traffic lights are critical for the SDV while it is not explicitly utilized in previous modules. We take it into account through a GRU network to refine the trajectory.
We initialize the hidden state with front camera features from the encoder module, and use each sample point in $\tau^*$ as the input.
%
It is demonstrated in Tab. \ref{tab:p2_ablation} that the front camera features indeed capture the information of traffic lights, which helps the SDV to launch or stop at intersections.

\subsection{Overall Loss for End-to-End Learning
}\label{sec: method-e2e learning}
We optimize our model with perception, prediction, planning in an end-to-end manner by exploiting the following loss functions:
\begin{equation}
    \mathcal{L} =  \mathcal{L}_{per} + \alpha \mathcal{L}_{pre} + \beta \mathcal{L}_{pla}.
\end{equation}
Note that we follow the protocol in~\cite{kendall2018multi,chen2018gradnorm} where the weights $\alpha$, $\beta$ are learnable rather than fixed, to balance the scale in different tasks according to gradients of the corresponding task loss.

\noindent\textbf{Perception Loss.}
This loss includes the segmentation loss for current and past frames, the mapping loss and an auxiliary depth loss.
For semantic segmentation, we use a top-k cross-entropy loss as in~\cite{hu2021fiery} since the BEV image is largely dominated by the background. For instance segmentation, the $l_2$ distance loss is adopted for the centerness supervision while $l_1$ distance loss for the offset and flow tasks.
We use a cross-entropy loss for the lane and drivable area prediction.
%
%
Current methods~\cite{philion2020lift,wang2021learning} utilize downstream tasks to implicitly optimize the depth prediction rather than a direct supervision, yet this approach is remarkably affected by the design of the final loss function  without clear explainability.
Therefore we generate the depth value with other networks beforehand, then use it to direct the prediction. More details are in the Appendix. 

\noindent\textbf{Prediction Loss.}
As our prediction module infers the future semantic segmentation and instance segmentation, we keep the same top-k cross-entropy loss as in Perception. Nonetheless, loss in future timestamps would be discounted exponentially due to the uncertainty of the prediction.

\noindent\textbf{Planning Loss.}
Our planning module first select the \textit{best} trajectory $\tau^*$ from the sampled trajectory set $\tau$ as in Eq. (\ref{eq:over_obj}), then a GRU-based network is applied to further refine the trajectory to obtain final output $\tau_o^*$.
Thus the planning loss contains two parts: a max-margin loss, which treats expert behavior $\tau_h$ as a positive example while trajectories sampled as negative ones, and a naive $l_1$ distance loss between the output and the expert trajectory. In particular, we set
\begin{equation}
    \mathcal{L}_{pla} = \max_{\tau} \left[ f(\tau_h, c) - f(\tau, c) + d(\tau_h, \tau) \right]_+ + d(\tau_h, \tau_o^* ),
\end{equation}
where $[ \cdot ]_+$ denotes ReLU and $d$ is the $l_1$ distance between input trajectories. 


\section{Experiments}\label{sec: experiments}
We evaluate ST-P3 in both open-loop and closed-loop environments. 
We adopt nuScenes dataset~\cite{caesar2020nuscenes} for the open-loop setting,  and CARLA simulator~\cite{dosovitskiy2017carla} for the closed-loop demonstration.
By default we take the $1.0s$ of past context and predict the future $2.0s$ contexts, corresponding to 3 frames in the past and 4 frames in the future. More details on protocols are provided in the Appendix.

\subsection{Open-loop Experimental Results on nuScenes}\label{sec: res-nuscenes}


\textbf{Perception.} We evaluate the models on map representation and semantic segmentation.
The perceived map includes drivable area and lanes - two most critical elements for the driving behavior, since the SDV is assumed to be driving in the drivable areas and keeping itself in the center of lanes.
The semantic segmentation focuses on vehicles and pedestrians, both of which are the main agents in a driving environment.
We use the Intersection-over-Union (IoU) as the metric, modeling the perception module as a BEV segmentation task.
As shown in Tab. \ref{tab:perception}, ST-P3 gets the highest mean value on the nuScenes validation set, surpassing previous SOTA by \textbf{2.51\%} with our Egocentric Aligned Accumulation algorithm.

\noindent\textbf{Prediction.} Predicting future segmentation in BEV is first proposed in~\cite{hu2021fiery}, thus we select it as our baseline.
We evaluate our model by the metric of IoU, existing panoptic quality (PQ), recognition quality (RQ), and segmentation quality (SQ) following metrics in video prediction area~\cite{kim2020video}.
Note that we predict vehicles which have been shown in the past frames only as we cannot predict those out of nowhere.
Results are shown in Tab. \ref{tab:prediction}. Our model achieves the state-of-the-art in \textit{all} metrics as a consequence of the novel design of the prediction module.
Though the Gaus. version performs a little worse than the Ber. one, we finally choose it for relatively smaller memory usage.



\noindent\textbf{Planning.} For the open-loop planning, we focus on two evaluation metrics: L2 error and collision rate, and adjust the planning horizon to $3.0s$ for a fair comparison. 
We compute the L2 error between the planned trajectory and the human driving trajectory, 
and evaluate how often the SDV would collide with other agents on the road.
%
%
Detailed comparison with previous methods is shown in Tab. \ref{tab:planning}. Note that the Vanilla algorithm is penalized based on how much the trajectory deviates from the ground truth, thus it achieves the lowest L2 error but largest collision rates in all prediction horizons.
ST-P3 obtains the lowest collision rate, implying the superior safety of our planned trajectory.
%

\begin{table}[tb!]
\caption{\textbf{Perception results.} We report the BEV segmentation IoU (\%) of intermediate representations and their mean value. ST-P3 outperforms in most cases}
\label{tab:perception}
\centering
\scalebox{0.8}{
\begin{tabular}{p{0.2\textwidth}|>{\centering}p{0.12\textwidth}>{\centering}p{0.12\textwidth}>{\centering}p{0.12\textwidth}>{\centering}p{0.12\textwidth}>{\centering\arraybackslash}p{0.12\textwidth}}
\toprule
\multirow{2}{*}{Method}  & \cellcolor{gray!30}Mean Value & Drivable Area     & \multirow{2}{*}{Lane}     & \multirow{2}{*}{Vehicle}     & \multirow{2}{*}{Pedestrian}          \\ \midrule
VED~\cite{lu2019monocular}      & \cellcolor{gray!30}28.19  & 60.82 & 16.74 & 23.28 & 11.93  \\
VPN~\cite{9123682}                       & \cellcolor{gray!30}30.36  & 65.97 & 17.05 & 28.17 & 10.26  \\
PON~\cite{roddick2020predicting}      & \cellcolor{gray!30}30.52  & 63.05 & 17.19 & 27.91 & 13.93  \\
Lift-Splat~\cite{philion2020lift} & \cellcolor{gray!30}34.61 & 72.23 & 19.98 & 31.22 & 15.02  \\
IVMP~\cite{wang2021learning}      & \cellcolor{gray!30}36.76 & 74.70  & 20.94 & 34.03 & \textbf{17.38}  \\
FIERY~\cite{hu2021fiery}    & \cellcolor{gray!30}40.18  & 71.97 & 33.58 & 38.00 & 17.15  \\ \midrule
\textbf{ST-P3} & \cellcolor{gray!30}\textbf{42.69} & \textbf{75.97} & \textbf{40.20} & \textbf{40.10} & 14.48  \\
\bottomrule
\end{tabular}}
\end{table}

\begin{table}[tb!]
\caption{\textbf{Prediction results.} We report semantic segmentation IoU (\%) and instance segmentation metrics from video prediction area. The \textit{static} method assumes all obstacles static in the prediction horizon. ``Ber.'' denotes modeling the future uncertainty as Bernoulli distribution, while ``Gaus.'' means Gaussian distribution. ST-P3 achieves best performance on all temporal segmentation metrics}
\label{tab:prediction}
\centering
\scalebox{0.8}{
\begin{tabular}{p{0.2\textwidth}|>{\centering}p{0.28\textwidth}|>{\centering}p{0.12\textwidth}>{\centering}p{0.12\textwidth}>{\centering\arraybackslash}p{0.12\textwidth}}
\toprule
\multicolumn{1}{l|}{\multirow{2}{*}{Method}} & Future Semantic Seg. & \multicolumn{3}{c}{Future Instance Seg.} \\
\multicolumn{1}{c|}{}                          & IoU $\uparrow$             & PQ $\uparrow$           & SQ $\uparrow$          & RQ $\uparrow$          \\ \midrule
Static                                         & 32.20            &    27.64          &   70.05          &   39.08          \\
FIERY~\cite{hu2021fiery}      & 37.00                                      & 30.20         & 70.20        & 42.90        \\ \midrule
\textbf{ST-P3 Gaus.}                                   & 38.63                                    &  31.72        & 70.15       & 45.22       \\
\textbf{ST-P3 Ber.}                                  & \textbf{38.87}                                     & \textbf{32.09}        & \textbf{70.39}       & \textbf{45.59}       \\
\bottomrule
\end{tabular}}
\end{table}

\begin{table}[tb!]
\centering
\caption{\textbf{Open-loop planning results.} \textit{vanilla} approach is supervised with ground truth trajectories only. ST-P3 achieves the lowest collision rate in all time intervals}
\label{tab:planning}
\scalebox{0.8}{
\begin{tabular}{p{0.2\textwidth}|>{\centering}p{0.1\textwidth}>{\centering}p{0.1\textwidth}>{\centering}p{0.1\textwidth}|>{\centering}p{0.1\textwidth}>{\centering}p{0.1\textwidth}>{\centering\arraybackslash}p{0.1\textwidth}}
\toprule
\multirow{2}{*}{Method}                     & \multicolumn{3}{c|}{L2 ($m$) $\downarrow$} & \multicolumn{3}{c}{Collision (\%) $\downarrow$} \\ \cline{2-7} 
                                             & 1s      & 2s     & 3s     & 1s        & 2s        & 3s        \\ \midrule
Vanilla                     & \textbf{0.50}    & \textbf{1.25}   & \textbf{2.80}   & 0.68      & 0.98      & 2.76      \\
NMP~\cite{zeng2019end}      & 0.61    & 1.44   & 3.18   & 0.66      & 0.90      & 2.34       \\
Freespace~\cite{hu2021safe} & 0.56    & 1.27   & 3.08   & 0.65      & 0.86      & 1.64      \\  \midrule
\textbf{ST-P3}                & 1.33    & 2.11   & 2.90   & \textbf{0.23}      & \textbf{0.62}      & \textbf{1.27}      \\ \bottomrule
\end{tabular}}
\end{table}

\begin{table}[tb!]
\caption{\textbf{Closed-loop simulation results.} ST-P3 outperforms vision-based baselines in all scenarios, and achieves better route completion performance in long-range tests compared to LiDAR-based method. $^*$: LiDAR-based method}
\label{tab:close-planning}
\centering
\scalebox{0.8}{
\begin{tabular}{p{0.20\textwidth}|>{\centering}p{0.16\textwidth}>{\centering}p{0.16\textwidth}|>{\centering}p{0.16\textwidth}>{\centering\arraybackslash}p{0.16\textwidth}}
\toprule
\multirow{2}{*}{Method} & \multicolumn{2}{c|}{Town05 Short} & \multicolumn{2}{c}{Town05 Long} \\ \cline{2-5} 
                        & DS $\uparrow$     & \multicolumn{1}{c|}{RC $\uparrow$}  & DS $\uparrow$             & RC $\uparrow$             \\ \midrule
CILRS~\cite{codevilla2019exploring} & 7.47   & 13.40                    & 3.68           & 7.19           \\
LBC~\cite{chen2020learning}       & 30.97  & 55.01                    & 7.05           & 32.09          \\
Transfuser$^*$~\cite{prakash2021multi}                     & 54.52     & 78.41                    & \textbf{33.15}          & 56.36          \\ \midrule
\textbf{ST-P3}                     & \textbf{55.14}       & \textbf{86.74}                         & 11.45               & \textbf{83.15}               \\ 
\bottomrule
\end{tabular}}
\end{table}

\subsection{Closed-loop Planning Results on CARLA Simulator}\label{sec: res-carla}


We conduct closed-loop experiments in CARLA simulator to demonstrate the applicability and robustness of ST-P3. It is far more challenging since the driving errors would stack up and lead to dangerous crashes. 
Following~\cite{prakash2021multi}, we adopt the Route Completion (RC) - the percentage of route distance completed, and the Driving Score (DS) - RC weighted by an penalty factor that accounts for collisions with pedestrians, vehicles, \textit{etc}.
Tab. \ref{tab:close-planning} shows the comparison with two camera-based algorithms and a LiDAR-based SOTA method.
ST-P3 outperforms the camera-based methods on all metrics and is comparable with the LiDAR-based method.
%
Tansfuser~\cite{prakash2021multi} has a higher driving score in long routes mainly due to the lower penalty resulting from the shorter traveling distance.
%
ST-P3 obtains impressive route completion performance which indicates the ability of recovering from collisions, with help of the front-view vision refinement.

\subsection{Ablation Study}\label{sec: exp-ablation}

\begin{table}[tb!]
\caption{\textbf{Ablation on nuScenes validation set.} Exp.1-3 explore the effectiveness of depth supervision (Depth) and egocentric aligned accumulation module (EAA.) for perception. Exp.4-6 is on prediction module, with ``Dual.'' representing Dual Modelling and ``LFA.'' indicating loss for all timestamps. Exp.7-9 show the superiority of combining the sampler (S.) and refinement (R.) units in planning. ``V.IoU'' is the IoU metric of vehicles. ``V.PQ'' is the panoptic quality of vehicles. ``Col.'' denotes the collision rate}
\label{tab:p2_ablation}
\centering
\scalebox{0.8}{
\begin{tabular}{p{0.06\textwidth}|>{\centering}p{0.1\textwidth}>{\centering}p{0.1\textwidth}>{\centering}p{0.1\textwidth}>{\centering}p{0.1\textwidth}>{\centering}p{0.1\textwidth}>{\centering}p{0.1\textwidth}|>{\centering}p{0.1\textwidth}>{\centering}p{0.1\textwidth}>{\centering}p{0.1\textwidth}>{\centering\arraybackslash}p{0.1\textwidth}}
\toprule
Exp. & EAA. & Depth & Dual. & LFA. & S. & R. & V. IoU $\uparrow$ & V. PQ $\uparrow$ & L2 $\downarrow$ & Col. $\downarrow$ \\ \midrule
1    &         &         &       &       &       &     &38.00   &   -   &  - &    - \\
2    &\ding{51}&         &       &       &       &     &38.79   &   -   & -  &   -  \\
3    &\ding{51}&\ding{51}&       &       &       &     &40.10   &   -   &  - &   -  \\ \midrule
4    &\ding{51}&\ding{51}&       &       &       &     &37.09   &28.63  &  - &   -  \\
5    &\ding{51}&\ding{51}&\ding{51}&       &       &     &38.16   & 31.35 &  - &   -    \\
6    &\ding{51}&\ding{51}&\ding{51}&\ding{51}&       &     &38.63   & 31.72 &  - &   - \\  \midrule
7    &\ding{51}&\ding{51}&\ding{51}&\ding{51}&\ding{51}&       & - & - &2.128   &  0.850     \\
8    &\ding{51}&\ding{51}&\ding{51}&\ding{51}&       &\ding{51}& - & - &2.321   &  1.089    \\ 
9    &\ding{51}&\ding{51}&\ding{51}&\ding{51}&\ding{51}&\ding{51}& - & - &1.890   &  0.513    \\ 
\bottomrule
\end{tabular}}
\end{table}

Tab. \ref{tab:p2_ablation} shows the effectiveness of different modules in ST-P3. We report the vehicle IoU and vehicle PQ for perception and prediction tasks, L2 and collision rate for planning evaluation.
For Exp.1-3 in Tab. \ref{tab:p2_ablation} we present the impact of depth supervision and Egocentric Aligned Accumulation (EAA.) in perception. Note that Exp.1 is identical to FIERY~\cite{hu2021fiery}.
Our module improves \textbf{0.79\%} by adopting EAA. algorithem (Exp.2), and supervising depth explicitly brings an improvement to \textbf{1.31\%} (Exp.3).
Exp.4-6 demonstrate the impact of Dual Modelling and corresponding training method - loss for all states (LFA.) on prediction task.
Since Dual Modelling considers both uncertainty and historical continuity, the correctness of past features plays a vital role in it.
As the results show, these two strategies improves \textbf{1.54\%} and \textbf{3.09\%} to V.IoU and V.PQ respectively.
Exp.7-9 is on the sampler and GRU refinement unit in planning.
A sampler without front-view vision refinement (Exp.7) or an implicit model without prior sampling knowledge (Exp.8) both get a high L2 error and collision rate. Our design remarkably improves the safety and accuracy of the planned trajectory.

\section{Conclusions}

In this paper, we have proposed an interpretable end-to-end vision-based framework for autonomous driving tasks. The motivation behind the improved design is to boost feature representations both in spatial and temporal domains. An egocentric aligned accumulation to aggregate features in 3D space and preserve geometry information is proposed; a dual pathway modelling to reason about the probabilistic character of semantic representations across frames is devised; a prior knowledge refinement unit to take into account road elements is introduced. Together with these improvements within the ST-P3 pipeline, we achieve impressive performance compared to previous state-of-the-arts.

\section*{Acknowledgments}
The project is partially supported by the Shanghai Committee of Science and Technology (Grant No. 21DZ1100100). This work is also supported in part by National Key Research and Development Program of China (2020AAA0107600), Shanghai Municipal Science and Technology Major Project (2021SHZDZX0102) and NSFC (61972250, 72061127003).

%
%
\bibliographystyle{splncs04}
\bibliography{egbib}

\clearpage
\appendix
\noindent{\Large \textbf{Appendix}}
\section{Implementation Details of the Network}
\subsection{Architecture for Perception and Prediction}

\subsubsection{Spatial-Temporal Perception.}
For nuScenes dataset, we first crop and resize the original image $\mathbb{R}^{3 \times 900 \times 1600}$ to $\mathbb{R}^{3 \times 224 \times 480}$ and take the past 3 frames to the model, denoted by $I_t^n \in \mathbb{R}^{3 \times 224 \times 480}$, where $t \in \{1,2,3\}, n \in \{1,\dots,6\}$.
We use EfficientNet-b4~\cite{tan2019efficientnet} as the backbone, obtaining features $f_i^k \in \mathbb{R}^{C \times H_e \times W_e}$ and depth estimation $d_i^k \in \mathbb{R}^{D \times H_e \times W_e}$ where $C = 64, D = 48, H_e = 28, W_e = 60$.
Note that the depth ranges from $2m$ to $50m$ with spacing $1m$.
After spreading the feature across the entire ray of space according to the predicted depth distribution, camera images from all angles are denoted by $u_i^k \in \mathbb{R}^{C \times D \times H_e \times W_e}$.
Then with the ego-motion matrix, features for all surround cameras and timestamps could be transformed to the coordinate system centered at the SDV at time $t$, resulting in ego-centric features $\{u_i^{'}\}$.
Finally the BEV features maps $b_i \in \mathbb{R}^{C \times H_b \times W_b}$ could be sum pooled from $\{u_i^{'}\}$ with $H_b =200, W_b = 200$.

In order to boost the features, we can integrate historical information to every frame through an accumulation method in the temporal fusion step.
Then these features are fed into a temporal network realized by 3D convolutions to better align temporal features.
In particular, in order to take different receptive fields into account, we apply different 3D kernel sizes on the time channel with $[(2,3,3),(1,3,3)]$ and a pyramid pooling with kernel size $(2,200,200)$.
We concatenate all the outputs to feed into the final compression convolution layer, getting the final output $x_{1 \sim t} \in \mathbb{R}^{C \times H_b \times W_b}$.
Note that the output features keep the same shape with the input after passing through the temporal network, and features from different timestamps are integrated.

\noindent
\subsubsection{Future Prediction.}
In our experiments, we model the future uncertainty by two different distributions: Gaussian and Bernoulli.
For Gaussian distribution, the present feature $x_t$ is passed through several residual block layers and an average pooling layer to get the hidden state $\mathbb{R}^{32 \times 1 \times 1}$.
Then a 2D convolution with kernel size $(1,1)$ fits the mean and log variance of the Gaussian with $R^{32} \times R^{32}$.
For Bernoulli distribution, since each grid of the spatial is $0-1$, we pass $x_t$ to several residual block layers and through a LogSigmoid to regress the probability in each grid.
For the prediction network, we utilize the convolutional Gated Recurrent Unit as our basic module, and each gate is realized by a 2D convolution with stride $3$.
We combine the predicted features through a trusting gate implemented by a 2D convolution which has 2 output channels, and use it as the weight to sum the two predicted features.
The mixed predicted features are then used as the hidden state for the ``uncertainty'' pathway 
and the input state for the ``historical'' pathway. 
Through this method, we could recursively predicts future states $(\hat{x}_{t+1},\dots, \hat{x}_{t+H})$.

\noindent
\subsubsection{Decoders for BEV Representations.}
All task heads share the same backbone considering the robustness of the decoder,  which is implemented by the first three layers of ResNet-18~\cite{he2016resnet} and three upsampling layers of factor 2 with skip connections.
Features now has 64 channels, and then are passed to different heads according to the task requirement. Each head is composed of two 2D convolutions with different output channels.
In particular, we set that the number of output channel for all semantic segmentation heads is 2. However, we need to predict offset, center, future flow for instance segmentation task,  thus they are set as 2, 1 and 2 respectively.

\subsection{Planning}
In this section we will give more detailed the scoring functions and the implementation of the refinement GRU network.

\noindent\textbf{Safety Cost.} The SDV should not collide with other objects on the road and need to consider their future motion when planning its trajectory.
For this purpose, we use the predicted occupancy map to penalize the trajectories that intersect with the occupied regions.
%
%
Formally, for trajectories $\tau$ at each timestamp $t$, we will penalize $\tau$ if the SDV polygon $g$ intersects with the grids which are occupied by other objects (with a safety margin indicated by parameter $\lambda$), denoted by the $o_g(\tau, t, \lambda)$. The safety cost related to objects collision is given by: 
\begin{equation}
    f_o(\tau, o) = \sum_t \sum_g o_g(\tau, t, 0) + o_g(\tau, t, \lambda) v(\tau, t),
\end{equation}
where the first term penalizes the intersection grids whereas the second term penalizes the high-velocity motion with uncertainty occupancy.

Moreover, SDV usually follows a leading vehicle and should keep a certain safe distance from it which mainly depends on the speed of the leading vehicle.
%
%
Since the HD map is unavailable and thus the leading vehicle along the center lane is unknown, instead we compute the occupancy grid in front of the SDV with a distance $L$ determined by the SDV velocity.
Hence if we are too close to the leading vehicle, the cost will be large that the trajectory will be evaluated as bad.
However it cannot take effect when it is in lane change manoeuvres since the leading vehicle is not directly in front of SDV.

The above objectives focus on moving objects, meanwhile the vehicle should stay in the middle of the lane line as well.
We can ensure this traffic safety with the perceived lanes in the first stage, with a cost function that is set as the distance to lane lines.

\noindent\textbf{Refinement.} Getting the selected trajectory $\tau^*$ according to the scoring functions, we then refine it through a GRU network. Specially, the hidden state of the GRU is the feature of the front camera from the encoder, and the input state is the concatenation of the current position, the position from the selected trajectory $\tau^*$ and the target point. Note that the initial current position is $(0,0)$. Doing the same process, we could recursively obtain the final refinement trajectory taking into consideration the front camera information and the target point information.

\section{Depth Supervision}

In this section, we introduce how the depth maps of nuScenes dataset are generated for explicit supervision. 

\begin{figure}[tb!]
    \centering
    \subfigure[]{
    \includegraphics[width=0.75\textwidth]{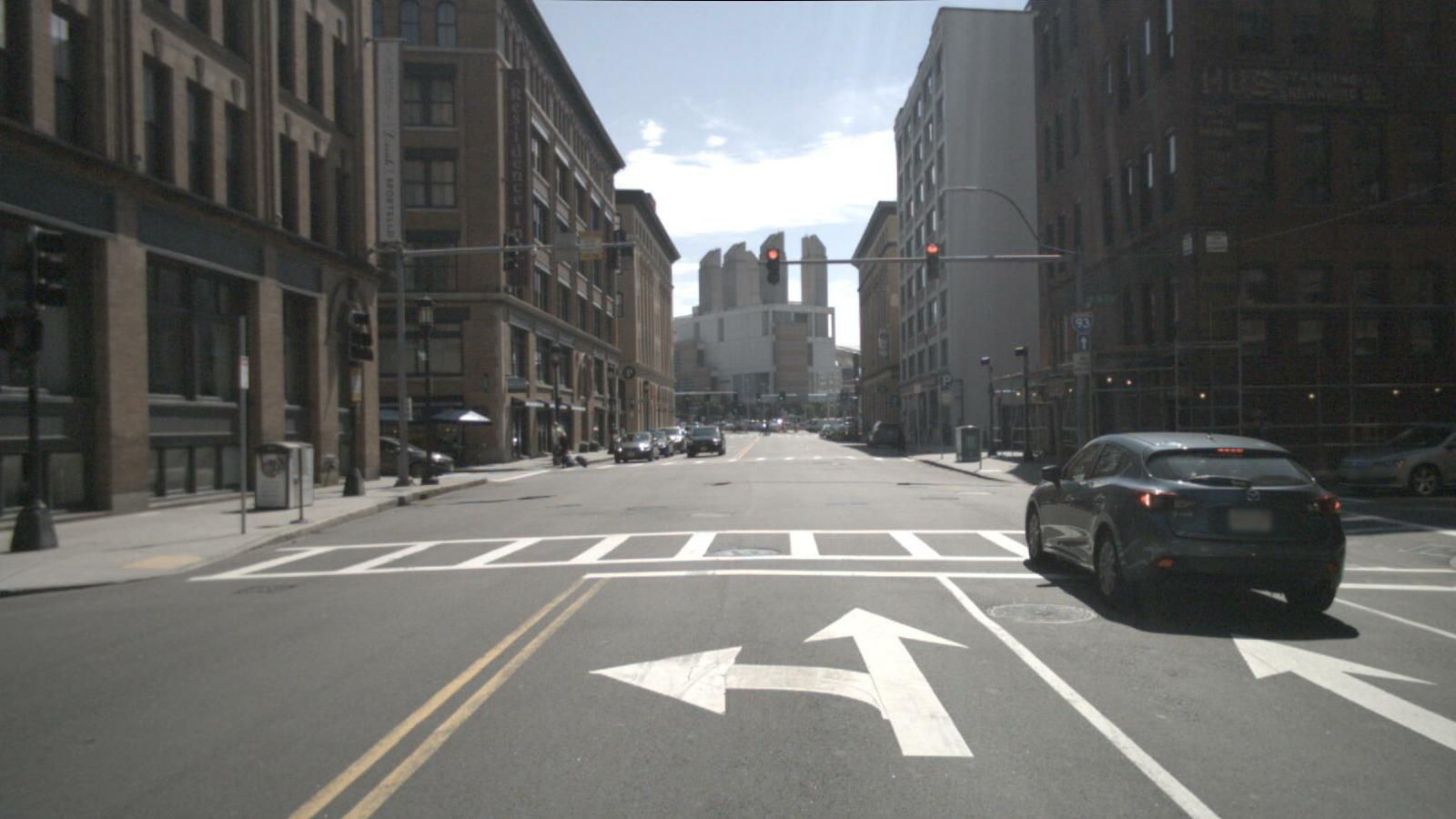}
    }
    \quad
    \subfigure[]{
    \includegraphics[width=0.75\textwidth]{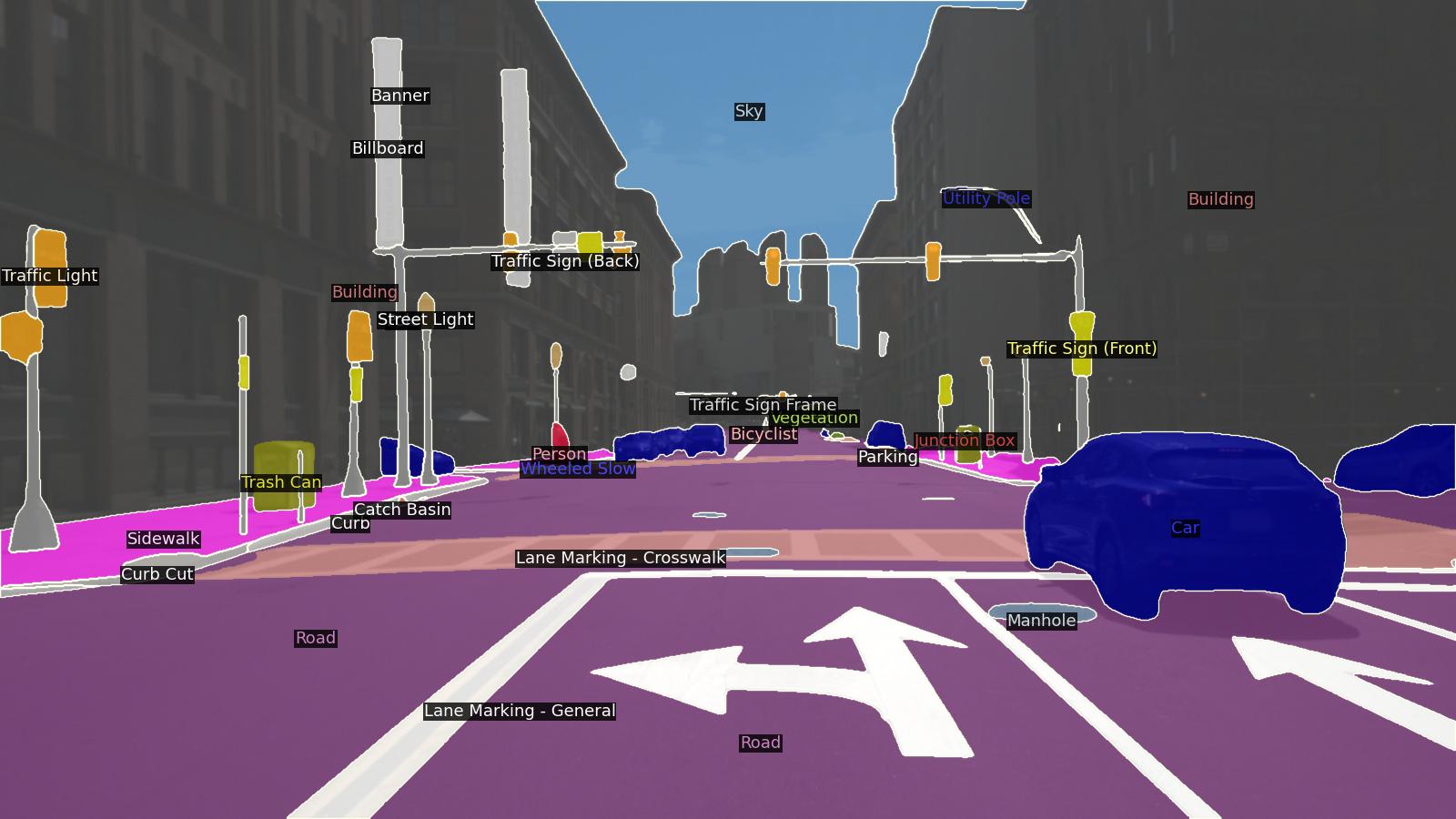}
    }
    \quad
    \subfigure[]{
    \includegraphics[width=0.75\textwidth]{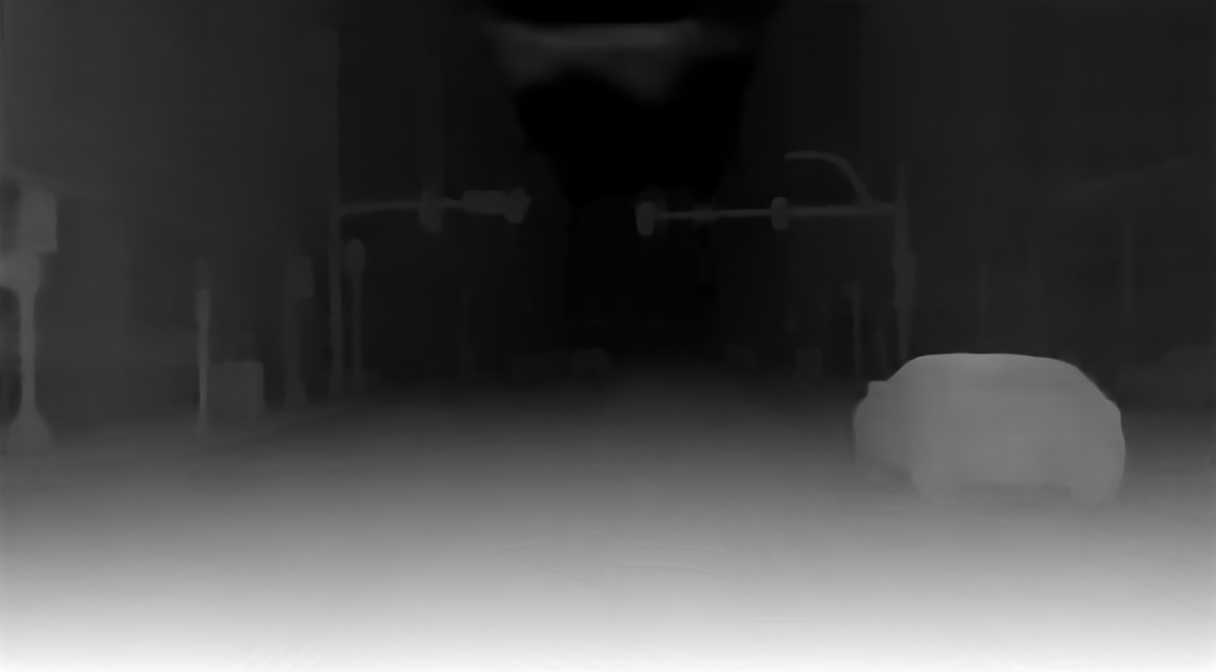}
    }
    \caption{(a) Original image; (b) Segmentation result; (c) Predicted depth map}
    \label{fig:depth_vis}
\end{figure}

We adopt a self-supervised, semantic segmentation required method named FSRE-Depth~\cite{Jung_2021_ICCV}, which has great performance on depth generation. Since FSRE-Depth use semantic segmentation as input, we first train a segmentation model on Mapillary Vistas Dataset~\cite{Mapillary}. Then we apply it on nuScenes dataset to generate semantic results, as shown in Fig. \ref{fig:depth_vis}(b).
In the next step, We train FSRE-Depth with front view images in nuScenes, using ResNet-50 as the backbone. The resolution of input images is $1216 \times 672$. After 45 epochs of training, the model can predict sufficient results on front view images. To leverage the performance of depth model, we further train it on each camera view for 15 epochs.
Thus, our depth maps of all camera views could be produced by this model individually. To evaluate our depth maps, we calculate depth estimation metrics with LiDAR points every 20 images. The result is shown in Tab. \ref{tab:depth_map_eval}, which indicates that our depth maps have decent quality.
This result in Fig. \ref{fig:depth_vis}(c) would be utilized to explicitly supervise the depth estimation in the perception module of ST-P3.


\begin{table}[]
\centering
\caption{Depth map evaluation. ARE: absolute relative error; SRE: square relative error; RSME: root mean square error; RSME log: root mean square logarithmic error; $\delta$: accuracy (threshold 1.25)}
\label{tab:depth_map_eval}
\scalebox{0.8}{
\begin{tabular}{p{0.2\textwidth}|>{\centering}p{0.13\textwidth}>{\centering}p{0.13\textwidth}>{\centering}p{0.13\textwidth}>{\centering}p{0.13\textwidth}>{\centering}p{0.13\textwidth}>{\centering}p{0.13\textwidth}>{\centering\arraybackslash}p{0.13\textwidth}}
\toprule
View        & ARE & SQE & RSME   & RSME log & $\delta$  & $\delta^2$  & $\delta^3$  \\ 
\midrule
FRONT       & 0.1522  & 3.5165 & 6.8756 & 0.2252   & 0.8720 & 0.9474 & 0.9701 \\
FRONT LEFT  & 0.2516  & 2.6591 & 5.7600 & 0.3035   & 0.7290 & 0.8755 & 0.9317 \\
FRONT RIGHT & 0.3030  & 5.9052 & 6.6903 & 0.3278   & 0.7067 & 0.8651 & 0.9231 \\
BACK        & 0.2297  & 5.5757 & 7.8486 & 0.3020   & 0.8011 & 0.9159 & 0.9535 \\
BACK LEFT   & 0.2752  & 3.6375 & 5.7691 & 0.3110   & 0.6982 & 0.8673 & 0.9279 \\
BACK RIGHT  & 0.3021  & 3.9396 & 6.2027 & 0.3413   & 0.6605 & 0.8503 & 0.9279 \\ \bottomrule
\end{tabular}}
\end{table}

\section{Experiments}
\subsection{Protocols}
\indent\textbf{Dataset.}
We evaluate ST-P3 in both open-loop and closed-loop environments. 
We adopt nuScenes dataset~\cite{caesar2020nuscenes} for the open-loop setting,  and CARLA simulator~\cite{dosovitskiy2017carla} for the closed-loop demonstration.

%
%
%
For nuScenes, by default we take the $1.0s$ of past context and predict the future $2.0s$ contexts, which corresponds to 3 frames in the past and 4 frames in the future.
Since each batch input to the model contains 7 frames of contexts, we just follow the offical split method to split the dataset into training, validation that consist of 26124 and 5719 samples, respectively.
%

%
For CARLA, 
we conduct closed-loop evaluation. Note that we still need to train our model in a open-loop manner first.
We follow the dataset collection in~\cite{prakash2021multi} and use the Town05 scenario for evaluation and the rest for training. 
%
%
%
%
It is worth mentioning that  the camera setup in CARLA only consists of 4 cameras; thus it could not cover the $360^{\circ}$ field of view around the SDV.
The context from the past 1.0$s$ and current 4 cameras images are passed to our model, then a trajectory is produced and executed by the simulator until SDV getting to the destination or surpassing the time limitation.

\noindent\textbf{Implementation Details.}
We adopt EfficientNet-B4~\cite{tan2019efficientnet} as the backbone, and detailed model description would be illustrated in the Supplementary. We use the Adam optimizer with a constant learning rate $2 \times 10^{-4}$. We train our model on 4 Tesla V100 GPUs for 20 epochs at mixed precision.
The BEV spatial is $(100m, 100m)$ around the SDV with resolution $(0.50m,0.50m)$ on nuScenes dataset, following the setting in~\cite{hu2021fiery} for a fair comparison. While on CARLA, it is $(40m, 40m)$ around SDV with resolution $(0.20m, 0.20m)$. The time interval between two consecutive frames is 0.5$s$ for both datasets.

\subsection{Open-loop Experimental Results on nuScenes}

\begin{figure}[tb!]
    \centering
    \subfigure{
    \includegraphics[width=0.98\textwidth]{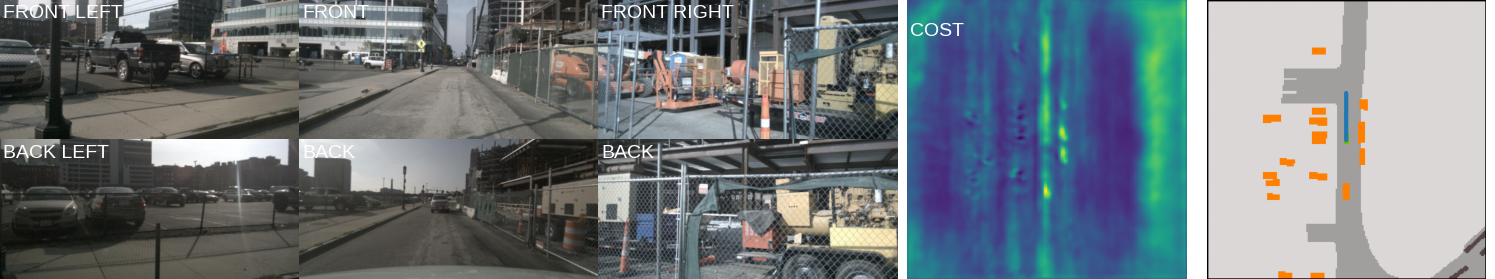}
    }
    \quad
    \subfigure{
    \includegraphics[width=0.98\textwidth]{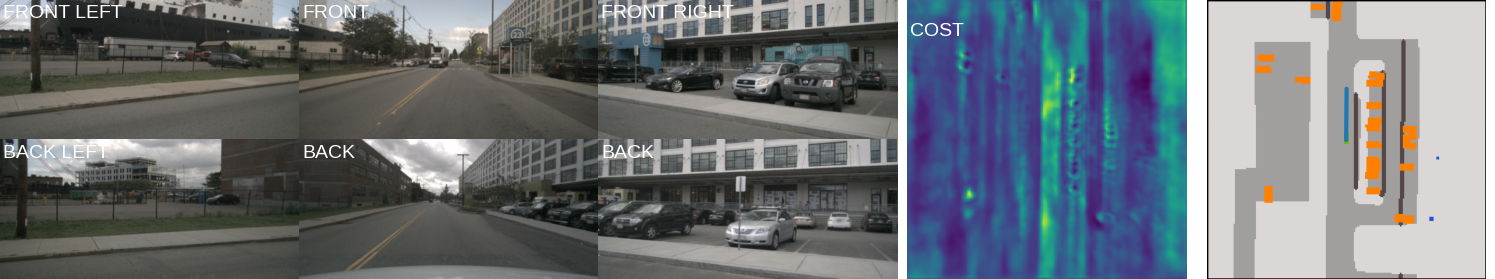}
    }
    \quad
    \subfigure{
    \includegraphics[width=0.98\textwidth]{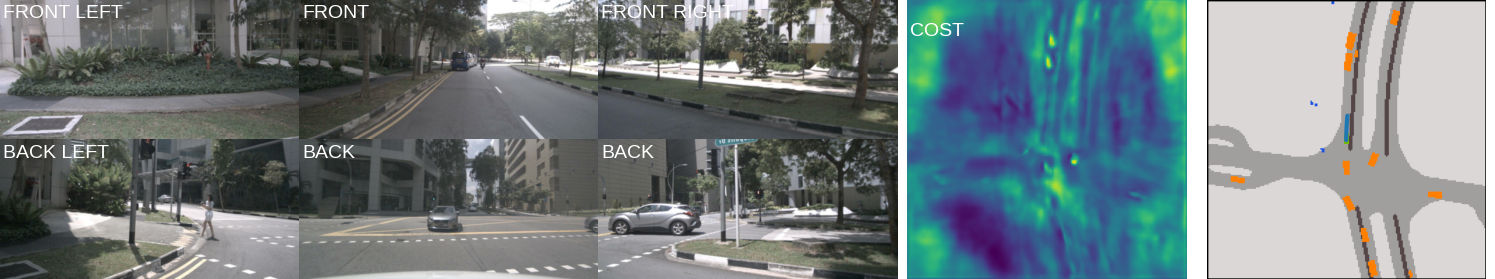}
    }
    \caption{Qualitative results of ST-P3 on the straight road. We show our BEV intermediate representations and planned trajectory (\textcolor{blue}{blue}) in the right. We also present the learned subcost map from prediction module. Note that a darker color indicates a smaller cost value}
    \label{fig:nu_plan}
\end{figure}

\begin{figure}[tb!]
    \centering
    \subfigure{
    \includegraphics[width=0.98\textwidth]{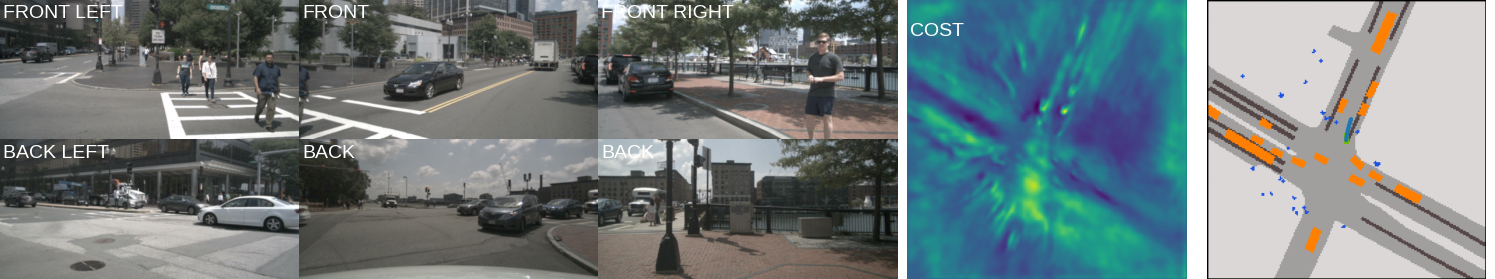}
    }
    \quad
    \subfigure{
    \includegraphics[width=0.98\textwidth]{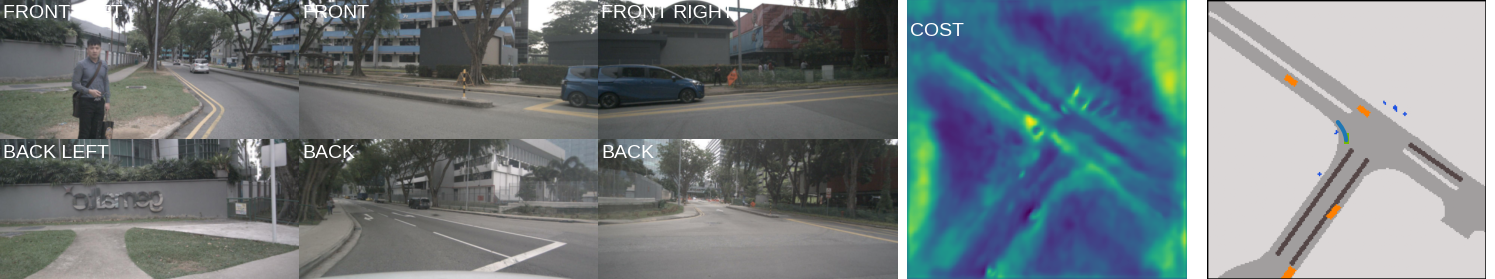}
    }
    \quad
    \subfigure{
    \includegraphics[width=0.98\textwidth]{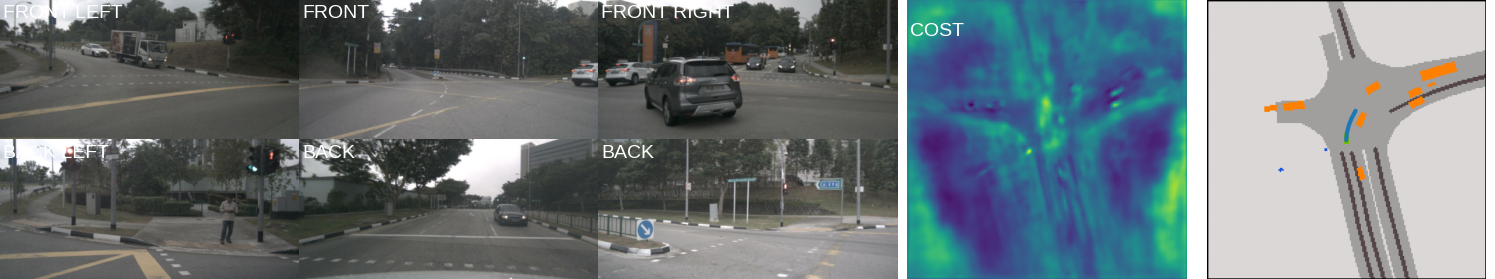}
    }
    \caption{Qualitative results of ST-P3 when turning at intersections}
    \label{fig:nu_turning}
\end{figure}

In this section we show more qualitative results on nuScenes dataset~\cite{caesar2020nuscenes}. We present the visualization of the learned cost volume and the composite graph of multiple semantic elements in the meantime.
Note that darker color means smaller cost value here, and vice versa. As shown in Fig. \ref{fig:nu_plan}-\ref{fig:nu_turning}, basically all the places occupied by cars have higher cost values, while open drivable areas have lower cost values.

\subsection{Closed-loop Planning Results on CARLA Simulator}
\begin{figure}[tb!]
    \centering
    \subfigure{
    \includegraphics[width=0.98\textwidth]{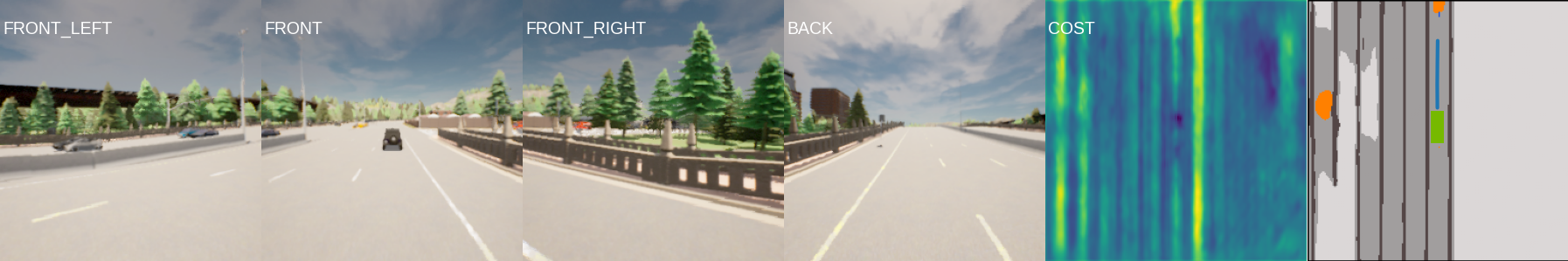}
    }
    \quad
    \subfigure{
    \includegraphics[width=0.98\textwidth]{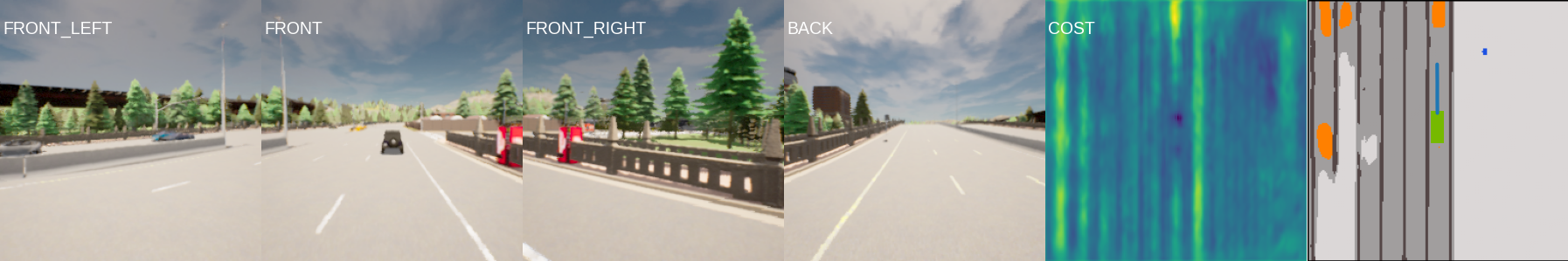}
    }
    \caption{Qualitative results of ST-P3 in closed-loop planning on the straight road}
    \label{fig:normal}
\end{figure}

\begin{figure}[tb!]
    \centering
    \subfigure{
    \includegraphics[width=0.98\textwidth]{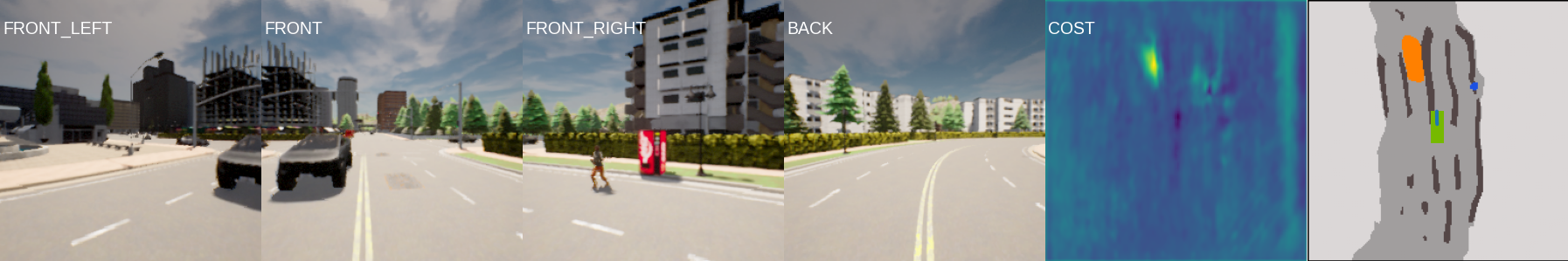}
    }
    \quad
    \subfigure{
    \includegraphics[width=0.98\textwidth]{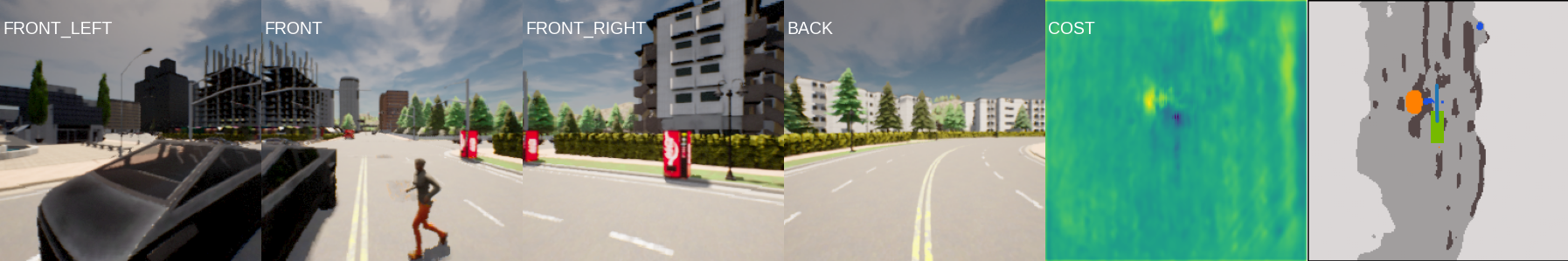}
    }
    \quad
    \subfigure{
    \includegraphics[width=0.98\textwidth]{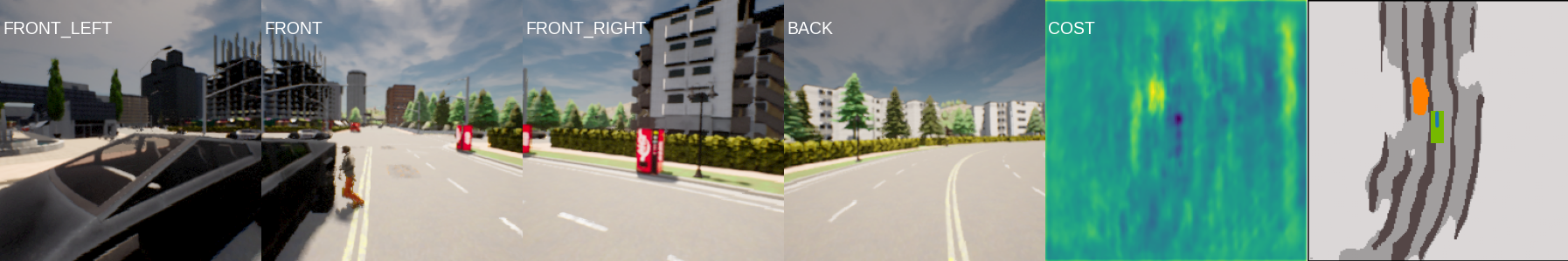}
    }
    \caption{ST-P3 predicts a slowing down trajectory when detecting a pedestrian}
    \label{fig:slow_down}
\end{figure}

\begin{figure}[tb!]
    \centering
    \subfigure{
    \includegraphics[width=0.98\textwidth]{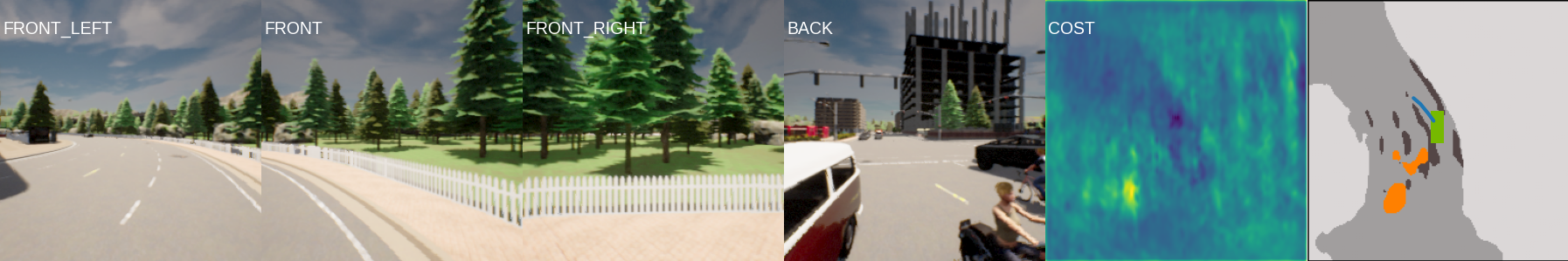}
    }
    \quad
    \subfigure{
    \includegraphics[width=0.98\textwidth]{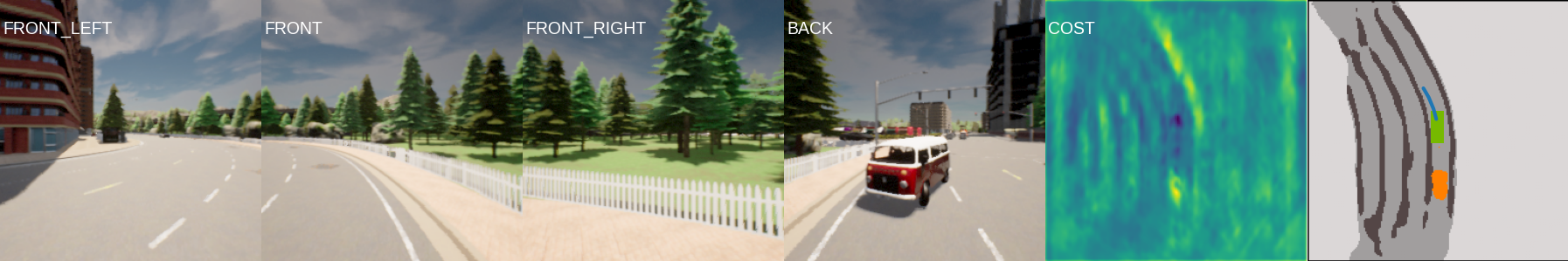}
    }
    \caption{BEV representations of ST-P3 become normal when driving on a predetermined trajectory (centerlines)}
    \label{fig:error_recover}
\end{figure}

\begin{figure}[tb!]
    \centering
    \subfigure{
    \includegraphics[width=0.98\textwidth]{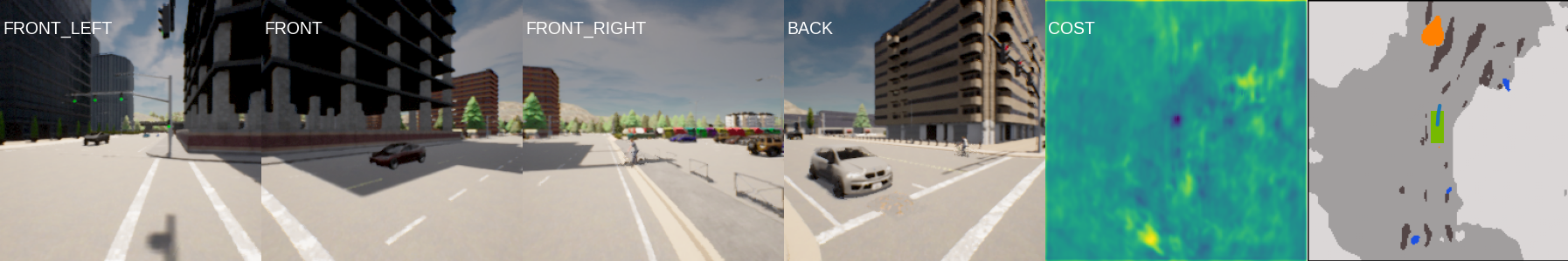}
    }
    \quad
    \subfigure{
    \includegraphics[width=0.98\textwidth]{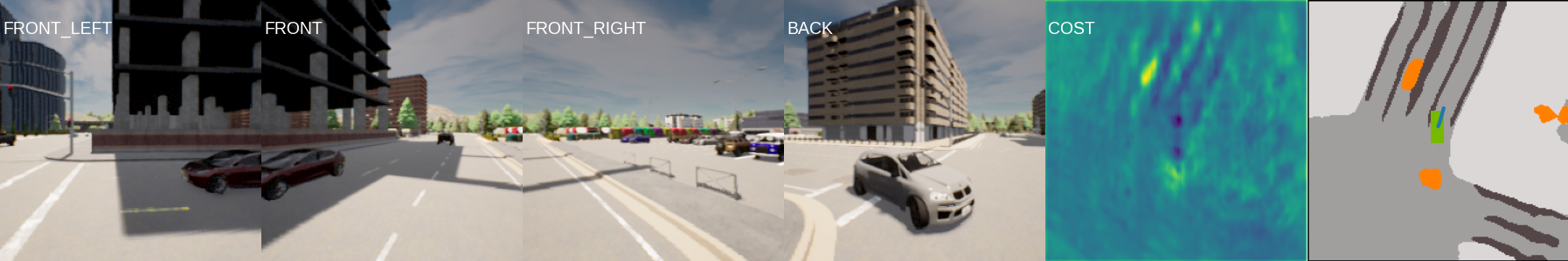}
    }
    \caption{Qualitative results of ST-P3 in closed-loop when turning at intersections}
    \label{fig:turning}
\end{figure}

In this section we show the qualitative results on the CARLA simulator~\cite{dosovitskiy2017carla}, similar to the open-loop results.
As shown in Fig. \ref{fig:error_recover}, when the car deviates from the center of lanes, the detection accuracy of the map drop significantly.
This is probably because the data collection strategy that most training data is on normal circumstance.
When the map accuracy is low, the traditional method which utilizes the sampler and cost map generally deviates from the expected track behavior, but due to our refinement operation, the car can still travel to the expected track correctly.
We provide more visualization in different scenarios in Fig. \ref{fig:normal}-\ref{fig:turning}.

\end{document}